\pdfoutput=1
\documentclass[]{dataflow}

\usepackage{latexsym}
\usepackage{textcomp}
\usepackage{amsmath}
\usepackage{array}
\usepackage{makecell}
\usepackage{float}
\usepackage{flafter}
\usepackage[ruled,vlined]{algorithm2e}
\usepackage{comment}
\usepackage{pifont}
\usepackage{fontawesome5}
\usepackage{url}
\newcommand{\cmark}{\ding{51}}
\newcommand{\xmark}{\ding{55}}
\graphicspath{{./}}
\makeatletter
\def\input@path{{./}}
\makeatother

\title{\textsc{MathDebugger}: A Benchmark for Type-Aware Error Detection in Synthetic Math Data}

\author[1,*]{Hao Liang}
\author[1,*]{Meiyi Qiang}
\author[2,*]{Yuying Li}
\author[3]{Zefeng He}
\author[4]{Xiaochen Ma}
\author[1]{Ruichuan An}
\author[5]{Yongzhen Guo}
\author[1]{Zhengzhou Zhu}
\author[1]{Bin Cui}
\author[1,\dagger]{Wentao Zhang}

\affiliation[1]{Peking University, Beijing, China}
\affiliation[2]{Tsinghua University, Beijing, China}
\affiliation[3]{Nanjing University, Nanjing, China}
\affiliation[4]{The Hong Kong University of Science and Technology, Hong Kong SAR, China}
\affiliation[5]{Ant Group, Beijing, China}

\contribution[*]{Equal contribution}
\contribution[\dagger]{Corresponding author}

\def\emailicon{\raisebox{-1.5pt}{\includegraphics[height=1.05em]{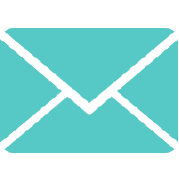}}}
\def\githubicon{\raisebox{-1.5pt}{\includegraphics[height=1.05em]{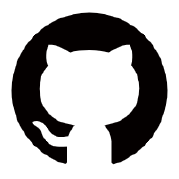}}}

\def\websiteicon{\raisebox{-0.5pt}{\faGlobe}}

\newcommand{\sourcelink}{https://github.com/haolpku/MathDebugger}
\newcommand{\websitelink}{https://haolpku.github.io/MathDebugger/}
\checkdata[\emailicon\hspace{0.3em} Correspondence]{\email{wentao.zhang@pku.edu.cn}}
\checkdata[\websiteicon\hspace{0.3em} Project Website]{\url{\websitelink}}
\checkdata[\githubicon\hspace{0.3em} Source Code and Dataset]{\url{\sourcelink}}

\abstract{Synthetic math data is widely used to scale LLM training, but its correctness is not guaranteed. We introduce \textsc{MathDebugger}, the first text-only, type-aware benchmark for jointly detecting \emph{and} classifying errors in synthetic mathematical questions and answers. \textsc{MathDebugger} contains 2{,}000 correct questions, 2{,}000 erroneous questions (balanced across four fine-grained error types), and 2{,}000 annotated answers (610 erroneous, across three error types). All items have correctness labels, and erroneous items are additionally assigned fine-grained error types; per-type Fleiss' $\kappa$ ranges from 0.69 to 0.91. On 14 SOTA LLMs (6 closed-source, 8 open-source, 1.5B--72B) and 3 process reward models, even advanced systems such as GPT-o3 and DeepSeek-R1 remain far from saturating the benchmark, particularly on fine-grained error-type detection. We further observe a consistent \emph{solving-vs-verifying} gap: math-tuned and long-chain-reasoning models often trail their general-purpose siblings on verification. Error-type annotations also provide actionable supervision for correction, with gains that are significant under paired bootstrap ($p<0.01$).}

\hypersetup{
  pdftitle={MathDebugger: A Benchmark for Type-Aware Error Detection in Synthetic Math Data},
  pdfauthor={Hao Liang, Meiyi Qiang, Yuying Li, Zefeng He, Xiaochen Ma, Ruichuan An, Yongzhen Guo, Zhengzhou Zhu, Bin Cui, Wentao Zhang}
}

\renewcommand{\today}{August 5, 2026}

\begin{document}
\maketitle

\renewcommand{\thefootnote}{\fnsymbol{footnote}}
\setcounter{footnote}{0}
\renewcommand{\thefootnote}{\arabic{footnote}}
\pagestyle{fancy}
\fancyhf{}
\fancyhead[L]{OpenDCAI Technical Report}
\fancyhead[R]{\thepage}

\newpage
\tableofcontents
\clearpage

\section{Introduction}
LLMs have demonstrated exceptional performance across a wide range of tasks~\cite{chatgpt, llama}. Prior work has established that data quality plays a crucial role in their success~\cite{yang2024qwen2, bai2024survey, abdin2024phi}. Recent efforts have explored data error detection techniques to improve LLM training efficiency and effectiveness~\cite{yan2024errorradar, ma2024large, he2025can}.
\begin{figure*}[htbp]
    \centering 
    \makebox[\textwidth]{\includegraphics[width=1.00\textwidth]{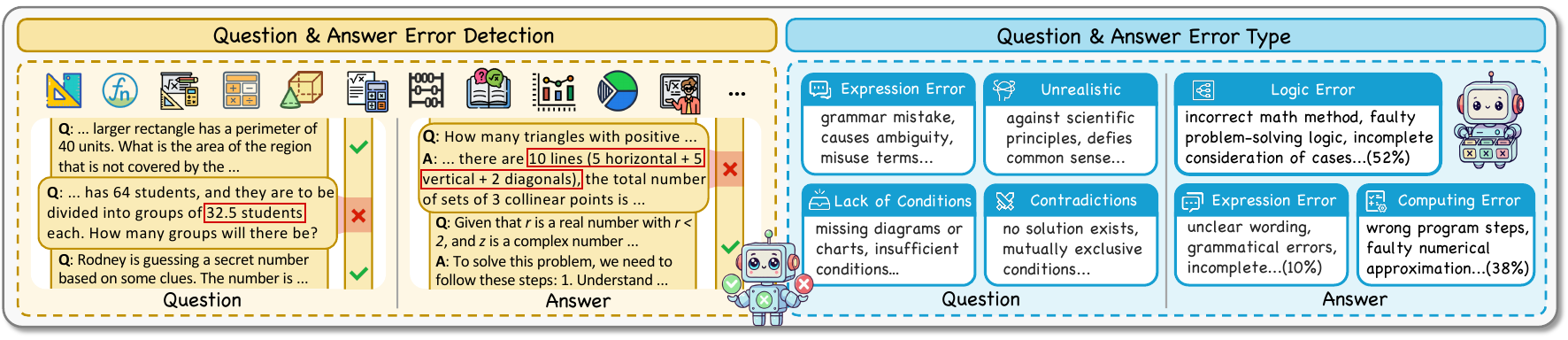}}
    \caption{Overview of the \textsc{MathDebugger} benchmark. \textsc{MathDebugger} poses two challenges: detecting errors in mathematical questions and answers, and identifying their error types.}
    \label{Fig.cover}
    \vspace{-2mm}
\end{figure*}

Among the various types of data used to train LLMs, mathematical data is especially valuable because it exercises a model's reasoning capabilities~\cite{yang2024qwen2, guo2025deepseek}. Collecting such data at scale is difficult: high-quality math QA pairs are scarce, so large corpora increasingly rely on synthesis~\cite{zhou2024jiuzhang3, toshniwal2024openmathinstruct}; the resulting data are not guaranteed to be correct~\cite{toshniwal2024openmathinstruct}; and automated correctness evaluation remains challenging~\cite{song2025prmbench,zheng2024processbench,zhou2024your}. Outcome and process reward models~\cite{zhang2024generative,skyworkopeno12024, wang2024math, song2025prmbench} and specialized step-verification benchmarks~\cite{song2025prmbench, zheng2024processbench, zhou2024your} have been proposed, but they focus on scoring reasoning \emph{steps}, not on auditing the training data itself. \textsc{MathDebugger} targets the complementary task of \emph{error detection in synthetic mathematical training data}, which we operationalize through two objectives:

\textbf{Question and Answer Error Detection.}
ErrorRadar~\cite{yan2024errorradar} primarily targets multimodal data, while UMP~\cite{ma2024large} focuses on reasoning over unreasonable questions. DeltaBench~\cite{he2025can} examines errors in long-chain CoT reasoning, such as step-level mistakes that may still yield correct answers. In contrast, none of these works directly assess the correctness of textual math question--answer pairs, which is the focus of our benchmark.

\textbf{Diverse Error Types.}
To rigorously assess whether LLMs truly understand the correctness of QA, it is essential to include diverse error types. A benchmark that categorizes errors can facilitate fine-grained evaluation and guide targeted improvements in synthetic data quality.

To address these challenges, we introduce \textsc{MathDebugger}, a benchmark designed to evaluate LLMs' ability to detect errors and identify error types in synthetic mathematical data. To construct the benchmark, we design 10 novel question-rewriting prompts for generating erroneous data and 16 rewriting prompts (11 more than those in MuggleMath~\cite{li2024mugglemathassessingimpactquery}) to enhance the diversity of synthesized questions. \textsc{MathDebugger} consists of 2,000 correct questions, 2,000 erroneous questions, and 2,000 annotated answers, each labelled for correctness and, when erroneous, assigned a fine-grained error type. We conduct large-scale experiments on \textsc{MathDebugger} with six widely used closed-source LLMs, eight open-source LLMs (1.5B--72B), and three state-of-the-art PRMs. Figure~\ref{Fig.cover} illustrates the benchmark's error-detection and error-type-identification tasks.

The core contributions of this work are:
\begin{itemize}
\item \textbf{\textsc{MathDebugger} Benchmark.} We introduce \textsc{MathDebugger}, the first text-only, type-aware benchmark covering both question- and answer-side errors with a fine-grained typed taxonomy.

\item \textbf{Diverse and Controlled Data Synthesis.} We propose 10 novel question-rewriting prompts for generating erroneous data and 16 rewriting prompts (substantially extending prior work) to enrich data diversity. All synthesized questions and answers are carefully validated by human annotators, with per-error-type Fleiss' $\kappa$ ranging from 0.69 to 0.91.

\item \textbf{Comprehensive Evaluation.} We benchmark 6 closed-source LLMs, 8 open-source LLMs (1.5B--72B), and 3 PRMs. Our results show that even strong current systems remain far from saturating \textsc{MathDebugger}, highlighting its difficulty and diagnostic value. We further provide a random baseline, multi-seed variance, bootstrap 95\% CIs, and prompt-format ablations.

\item \textbf{Error-Type-Aware Correction.} We empirically demonstrate that fine-grained error-type annotations not only support diagnostic evaluation, but also provide actionable supervision for correcting erroneous mathematical questions and answers, consistently improving correction accuracy across models and error categories.
\end{itemize}

\section{Related Work}
\subsection{Mathematical Benchmarks}
Recent work has substantially advanced benchmarks for mathematical reasoning. GSM8K~\cite{cobbe2021training} and MATH~\cite{hendrycks2021measuring} are foundational benchmarks for grade-school arithmetic and competition-level mathematics, respectively. SuperCLUE-Math~\cite{xu2024superclue}, MathBench~\cite{liu2024mathbench}, Omni-MATH~\cite{gao2024omni}, and FrontierMath~\cite{glazer2024frontiermath} extend this landscape through broader language coverage, mathematical domains, difficulty ranges, and stronger controls for data contamination. PRMBench~\cite{song2025prmbench} and ProcessBench~\cite{zheng2024processbench} evaluate step-level verification, while MathTrap~\cite{zhao2024exploring} focuses on logical traps in reasoning chains. A gap nevertheless remains in evaluating whether LLMs can identify low-quality or incorrect mathematical training data, a capability essential to data filtering and cleaning.

\begin{table*}[t]
\centering
\small
\resizebox{0.85\linewidth}{!}{
\begin{tabular}{lccccc}
\toprule
\textbf{Benchmark} & \textbf{Granularity} & \textbf{Q-err} & \textbf{A-err} & \textbf{Typed} & \textbf{Size} \\
\midrule
ProcessBench~\cite{zheng2024processbench} & step       & \xmark & \cmark & \xmark & 3{,}400 \\
PRMBench~\cite{song2025prmbench}          & step       & \xmark & \cmark & partial & 6{,}216 \\
ErrorRadar~\cite{yan2024errorradar}       & solution (multimodal) & \xmark & \cmark & \cmark & 2{,}500 \\
DeltaBench~\cite{he2025can}               & CoT step   & \xmark & \cmark & \xmark & 1{,}236 \\
UMP~\cite{ma2024large}                    & question (unreasonable) & \cmark & \xmark & \xmark & 582 \\
MATH-Minos~\cite{gao2024llm}              & step       & \xmark & \cmark & \xmark & -- \\
\midrule
\textbf{\textsc{MathDebugger}} (ours)     & \textbf{question + solution (text)} & \cmark & \cmark & \cmark & \textbf{6{,}000} \\
\bottomrule
\end{tabular}
}
\caption{Positioning of \textsc{MathDebugger} relative to closely related error / step-verification benchmarks. \textbf{Q-err}: errors in the question statement; \textbf{A-err}: errors in the answer/solution; \textbf{Typed}: fine-grained error-type labels.}
\label{tab:benchmark_comparison}
\end{table*}

Table~\ref{tab:benchmark_comparison} summarizes how \textsc{MathDebugger} differs from the closest existing benchmarks. ProcessBench and PRMBench operate at the \emph{reasoning-step} level and assume the underlying question is valid; they cannot surface errors in the problem statement itself. ErrorRadar targets multimodal math problems and provides typed errors only on the answer side. DeltaBench focuses on step-level flaws in long CoT traces where the \emph{final answer may still be correct}. UMP evaluates whether models refuse unreasonable questions but does not provide typed error labels and is limited to 582 items. MATH-Minos introduces step-wise natural-language feedback but does not provide a fine-grained error taxonomy. In contrast, \textsc{MathDebugger} is the first text-only benchmark that (i) covers \emph{both} question-side and answer-side errors, (ii) provides a fine-grained four-class (question) and three-class (answer) taxonomy, and (iii) reports per-error-type inter-annotator agreement.
\subsection{Mathematical Data Synthesis}
The demand for high-quality LLM training data has driven rapid progress in data synthesis. Existing approaches can be broadly divided into two categories. The first uses \textbf{LLM-based distillation}, with prompt engineering for synthesis tasks such as question rewriting. Representative examples include MetaMath~\cite{yu2023metamath}, KPDDS~\cite{huang2024key}, and JiuZhang3.0~\cite{zhou2024jiuzhang3}. The second uses \textbf{Monte Carlo Tree Search (MCTS)-based synthesis}, in which reward models guide search over candidate generations. Representative systems include ReST-MCTS*~\cite{zhang2024rest}, LLaMA-Berry~\cite{zhang2024llama}, and Mulberry~\cite{yao2024mulberry}.
We primarily adopt the first approach. The synthesized data undergo multiple rounds of manual annotation to form \textsc{MathDebugger}, providing a high-quality benchmark for auditing mathematical data.
\section{\textsc{MathDebugger} Benchmark}

\begin{figure*}[h]
    \centering
    \includegraphics[width=1.00\textwidth]{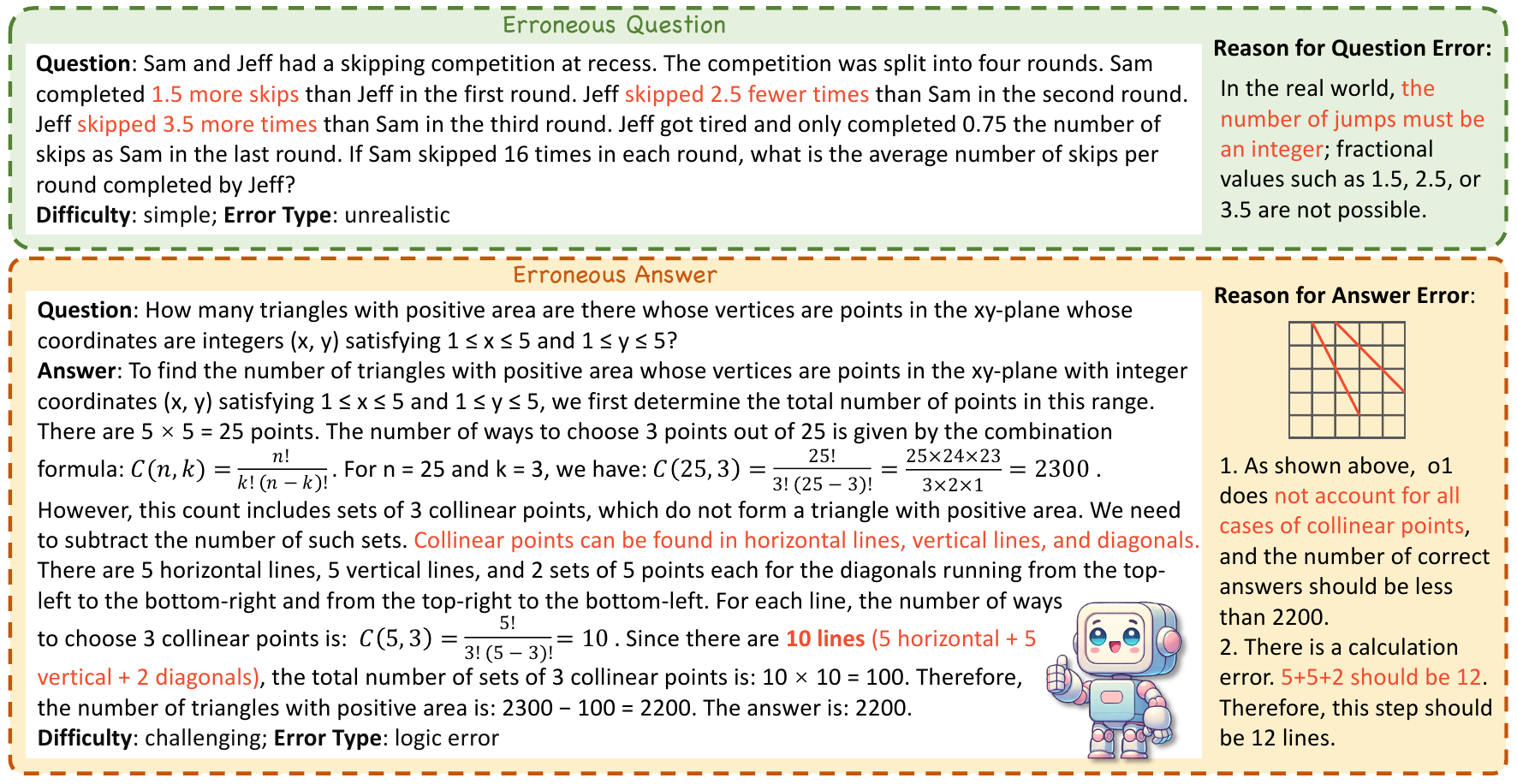}
    \caption{Two erroneous examples from \textsc{MathDebugger}: a SIMPLE question with an \emph{unrealistic} error, and a CHALLENGING question paired with an answer exhibiting a \emph{logic} error.}
    \label{Fig.example}
    \vspace{-2mm}
\end{figure*}

\subsection{Wrong Question Dataset Construction}
\label{sec:question_construction}
We use GSM8K and MATH as seed data, generating math questions with various error types using Qwen2.5-72B-Instruct, LLaMA-3.1-70B-Instruct, and GPT-4o. GSM8K-based outputs are classified as \textbf{SIMPLE} and MATH-based outputs as \textbf{CHALLENGING}. Mathematical experts analyse classic LLM errors and design prompts that guide the generators to produce target error types.

Many synthetic questions exhibit multiple error types; to ensure diversity, we first design ten specific erroneous conditions (Appendix Figure~\ref{Fig.error_question_prompt}), then consolidate them into four primary error types: (1) \textbf{Expression Error}, covering linguistic issues such as grammatical errors, ambiguous references, or redundant content; (2) \textbf{Lack of Conditions}, where information required for solution is missing; (3) \textbf{Contradictions}, where stated conditions are mutually inconsistent; (4) \textbf{Unrealistic}, where solutions violate common sense or real-world constraints. Under our single-label policy, annotators fully solve the item and label the earliest point along a mathematically valid reasoning path that invalidates the question or solution; later errors caused by that first violation are not assigned separate labels. Full definitions and the multiple-path/branched-reasoning rules are in Appendix~\ref{app:taxonomy}.

\paragraph{Annotation.} Questions are labelled by 30 recruited STEM graduate annotators and 10 reviewers from top-tier universities, who pass a multi-stage qualification screening. Annotators (i) solve the question, (ii) flag correctness, and (iii) for incorrect items, select the specific error type and provide a written justification. Every label undergoes dual independent review; annotations from an annotator are excluded if their consensus-adjudicated accuracy falls below $90\%$. The two-month effort cost $\sim\$50\text{k}$. Further details of the annotation protocol are in Appendix~\ref{app:annotation_protocol}.

\newcommand{\questionresultstable}{%
\begin{table}[htbp]
\centering
\renewcommand{\arraystretch}{1.06}
\begin{tabular*}{\textwidth}{@{\extracolsep{\fill}}lrrrrrrrr@{}}
\toprule
\multirow{3}{*}{\textbf{Model}} & \multicolumn{4}{c}{\textbf{Error detection}} & \multicolumn{4}{c}{\textbf{Error-type classification}} \\
\cmidrule(lr){2-5}\cmidrule(lr){6-9}
& \multicolumn{2}{c}{\textbf{Simple}} & \multicolumn{2}{c}{\textbf{Challenging}} & \multicolumn{2}{c}{\textbf{Simple}} & \multicolumn{2}{c}{\textbf{Challenging}} \\
\cmidrule(lr){2-3}\cmidrule(lr){4-5}\cmidrule(lr){6-7}\cmidrule(lr){8-9}
& Acc & F1 & Acc & F1 & Acc & M-F1 & Acc & M-F1 \\
\midrule
Random & 50.00 & 50.00 & 50.00 & 50.00 & 25.00 & 25.00 & 25.00 & 25.00 \\
\midrule
\multicolumn{9}{@{}l}{\textbf{Closed-source models}} \\
GPT-4o & 72.44 & 76.38 & 72.50 & 75.26 & \textbf{79.13} & \textbf{79.14} & 74.00 & 73.90 \\
GPT-o1 & 73.19 & 75.39 & \textbf{76.38} & 77.91 & 76.38 & 76.14 & 72.83 & 72.38 \\
GPT-o3 & 72.62 & 74.14 & 74.08 & 75.28 & 75.50 & 75.32 & 71.67 & 71.34 \\
GPT-o4-mini & 74.81 & 76.88 & 76.17 & 77.92 & 75.88 & 75.64 & 72.58 & 72.43 \\
Claude-3.5 & \textbf{76.31} & \textbf{78.33} & 75.91 & \textbf{78.25} & 74.88 & 74.90 & 70.67 & 70.85 \\
Gemini-2.5 & 71.81 & 70.77 & 62.54 & 55.87 & 64.50 & 63.05 & 62.33 & 61.12 \\
\midrule
\multicolumn{9}{@{}l}{\textbf{Open-source models}} \\
LLaMA-3.1-8B & 59.62 & 46.96 & 64.88 & 58.82 & 56.00 & 55.08 & 56.67 & 55.81 \\
LLaMA-3.3-70B & 71.37 & 76.42 & 69.71 & 74.01 & 75.38 & 75.24 & \textbf{77.83} & \textbf{77.90} \\
Qwen2.5-1.5B & 54.44 & 56.11 & 57.83 & 57.90 & 35.88 & 35.65 & 39.58 & 39.47 \\
Qwen2.5-7B & 67.75 & 74.02 & 71.33 & 76.13 & 55.13 & 55.62 & 58.75 & 59.17 \\
Qwen2.5-72B & 72.31 & 77.22 & 73.83 & 78.15 & 77.62 & 77.58 & 73.25 & 73.39 \\
Qwen2.5-Math-72B & 68.00 & 74.01 & 68.87 & 73.78 & 67.37 & 67.35 & 65.58 & 65.36 \\
QwQ-32B & 57.81 & 40.53 & 55.50 & 34.88 & 67.00 & 66.85 & 64.00 & 63.78 \\
DeepSeek-R1 & 71.25 & 72.46 & 74.50 & 75.94 & 70.37 & 70.21 & 70.75 & 70.50 \\
\bottomrule
\end{tabular*}
\caption{Question-side results (\%). M-F1 denotes Macro-F1. Best results in each task and split are bold. Multi-seed uncertainty is reported in Table~\ref{tab:planc_variance}.}
\label{tab:check_correct_questions}
\end{table}
}

\subsection{Wrong Answer Dataset Construction}
\label{sec:answer_construction}
To diversify the answer dataset, we (i) synthesize high-variation questions with 16 rewriting prompts (Appendix Figure~\ref{Fig.question_extension_prompt}). Since all questions in \textsc{MathDebugger} come from new synthesis, this also mitigates contamination. We then (ii) regenerate answers using models of varying capability: Qwen2.5-72B-Instruct (high-capacity), LLaMA-3.1-8B-Instruct (medium), and Qwen2.5-Math-1.5B-Instruct (lightweight), with both CoT and PoT prompting.

Answers are then annotated by the same pool of qualified annotators following a three-step protocol (check the question, check the answer, then label and justify the error type). Answer errors are categorized into three types: \textbf{Expression}, \textbf{Logic}, and \textbf{Computational} errors (definitions in Appendix~\ref{app:taxonomy}). After decontamination, we retain only items whose question is correct and whose answer exhibits at most one typical error type. Figure~\ref{Fig.example} shows representative erroneous instances.

\begin{table}[ht]
\centering
\begin{subtable}[t]{0.53\linewidth}
\centering
\begin{tabular}{@{}lrrr@{}}
\toprule
\textbf{Dataset} & \textbf{Error} & \textbf{Correct} & \textbf{Total} \\
\midrule
Simple      & 163 & 637  & 800  \\
Challenging & 447 & 753  & 1200 \\
Both        & 610 & 1390 & 2000 \\
\bottomrule
\end{tabular}
\caption{Answer dataset statistics}
\end{subtable}\hfill
\begin{subtable}[t]{0.39\linewidth}
\centering
\begin{tabular}{@{}lrr@{}}
\toprule
\textbf{Correctness} & \textbf{$N$} & \textbf{$\kappa$} \\
\midrule
Question--S & 1600 & \textbf{0.88} \\
Question--C & 2400 & \textbf{0.72} \\
Answer--S   & 800  & \textbf{0.87} \\
Answer--C   & 1200 & \textbf{0.76} \\
\bottomrule
\end{tabular}
\caption{Overall agreement}
\label{tab:overall_iaa}
\end{subtable}
\caption{Answer-dataset composition and overall inter-annotator agreement. Agreement is measured by Fleiss' $\kappa$; S and C denote SIMPLE and CHALLENGING.}
\label{tab:statistics_of_answer_dataset}
\end{table}

\subsection{Annotation Quality and Inter-Annotator Agreement}
\label{sec:iaa}
All questions and answers were independently annotated by two trained annotators, followed by expert review. We report Fleiss' $\kappa$ at both the correctness level and the fine-grained error-type level.

Table~\ref{tab:overall_iaa} reports overall agreement, while Appendix Table~\ref{tab:per_type_iaa_appendix} gives the question- and answer-type breakdowns. Binary correctness annotations reach almost perfect agreement on SIMPLE ($\kappa \geq 0.87$) and substantial agreement on CHALLENGING ($\kappa \geq 0.72$). Question-type $\kappa$ ranges from $0.88$ (\textit{lack of conditions} on SIMPLE) to $0.69$ (\textit{contradictions} on CHALLENGING), while answer-type $\kappa$ ranges from $0.70$ to $0.91$. We retain the lowest values as measured; they reflect genuine semantic boundary ambiguity that we did not smooth over, and all residual disagreements were adjudicated by domain experts rather than re-labelled to inflate agreement.

\subsection{Statistics of \textsc{MathDebugger}}
The question split comprises 2,000 correct and 2,000 incorrect questions; the incorrect half is balanced across the four error types (500 per class), split into 200 SIMPLE (GSM8K-seeded) and 300 CHALLENGING (MATH-seeded) per class. The answer split pairs 2,000 correct questions with their answers (40\% SIMPLE, 60\% CHALLENGING; 30.5\% of answers are incorrect). Error counts over the 610 erroneous answers are \textit{logic}=317 (52\%), \textit{computing}=232 (38\%), and \textit{expression}=61 (10\%), consistent with the distributions reported by MATH-Minos~\cite{gao2024llm} and ErrorRadar~\cite{yan2024errorradar}. Table~\ref{tab:statistics_of_answer_dataset} summarizes the answer split, while Appendix Figures~\ref{Fig.question_dataset}--\ref{Fig.answer_dataset} provide detailed composition charts by difficulty and error type.

\section{Experiments}

\subsection{Experimental Setting}
We evaluate four sub-tasks: question error detection, answer error detection, question error-type detection, and answer error-type detection (Q-Detect, A-Detect, Q-Type, and A-Type). The first two are binary; the latter are 4-way and 3-way classification tasks. Binary F1 in the model tables treats the \emph{correct/no-error} class as positive, matching the released evaluation mapping. Because this convention can obscure errors on the imbalanced answer split, Appendix~\ref{app:evaluation_protocol} additionally reports error-class F1 and Macro-F1. Type classification uses Accuracy and Macro-F1. The appendix also specifies the prompts, parsing rules, and treatment of unparseable outputs.

\subsection{Models}
\label{sec:models}
We evaluate 14 LLMs and three PRMs. \textbf{Closed-source LLMs} comprise GPT-4o, o1, o3, o4-mini~\cite{achiam2023gpt,chatgpt}, Claude-3-5-Sonnet, and Gemini-2.5-Flash~\cite{team2023gemini}. \textbf{Open-source LLMs} comprise LLaMA-3.1/3.3 (8B--70B)~\cite{dubey2024llama}, Qwen2.5-Instruct and Qwen2.5-Math (1.5B--72B)~\cite{yang2024qwen2}, QwQ-32B-Preview~\cite{qwq-32b-preview}, and DeepSeek-R1~\cite{guo2025deepseek}. The \textbf{PRMs} are Math-Shepherd-7B~\cite{wang2024math}, Skywork-PRM-7B~\cite{skyworkopeno12024}, and Qwen2.5-Math-PRM-7B~\cite{zhang2025lessons}. Following~\citet{cui2025process}, we evaluate PRMs only on answer detection by treating the full reasoning trace as one step and thresholding its score.

\questionresultstable
\FloatBarrier

\newcommand{\answerresultstable}{%
\begin{table}[htbp]
\centering
\renewcommand{\arraystretch}{1.04}
\begin{tabular*}{\textwidth}{@{\extracolsep{\fill}}lrrrrrrrr@{}}
\toprule
\multirow{3}{*}{\textbf{Model}} & \multicolumn{4}{c}{\textbf{Error detection}} & \multicolumn{4}{c}{\textbf{Error-type classification}} \\
\cmidrule(lr){2-5}\cmidrule(lr){6-9}
& \multicolumn{2}{c}{\textbf{Simple}} & \multicolumn{2}{c}{\textbf{Challenging}} & \multicolumn{2}{c}{\textbf{Simple}} & \multicolumn{2}{c}{\textbf{Challenging}} \\
\cmidrule(lr){2-3}\cmidrule(lr){4-5}\cmidrule(lr){6-7}\cmidrule(lr){8-9}
& Acc & F1 & Acc & F1 & Acc & M-F1 & Acc & M-F1 \\
\midrule
Random & 50.00 & 50.00 & 50.00 & 50.00 & 33.33 & 33.33 & 33.33 & 33.33 \\
\midrule
\multicolumn{9}{@{}l}{\textbf{Closed-source models}} \\
GPT-4o & 80.75 & 88.02 & 77.25 & \textbf{82.97} & 59.51 & 46.93 & 57.49 & 42.11 \\
GPT-o1 & 76.62 & 84.66 & \textbf{78.58} & 82.24 & 62.58 & 48.17 & 60.40 & 41.52 \\
GPT-o3 & 72.00 & 80.72 & 74.50 & 77.99 & \textbf{64.42} & \textbf{55.05} & \textbf{63.31} & 50.02 \\
GPT-o4-mini & 78.12 & 85.67 & 78.08 & 81.67 & 61.96 & 52.80 & 62.42 & \textbf{52.77} \\
Claude-3.5 & 73.12 & 81.98 & 73.83 & 78.84 & 60.12 & 48.17 & 56.38 & 38.21 \\
Gemini-2.5 & 76.25 & 85.13 & 68.25 & 75.18 & 61.96 & 49.48 & 60.85 & 45.46 \\
\midrule
\multicolumn{9}{@{}l}{\textbf{Open-source models}} \\
LLaMA-3.1-8B & 63.25 & 73.80 & 62.83 & 67.25 & 45.40 & 40.39 & 49.22 & 36.40 \\
LLaMA-3.3-70B & 79.25 & 86.99 & 76.17 & 82.01 & 57.67 & 51.28 & 58.61 & 46.84 \\
Qwen2.5-1.5B & 67.25 & 78.45 & 58.42 & 68.28 & 49.69 & 36.05 & 45.41 & 33.19 \\
Qwen2.5-7B & 81.50 & 88.91 & 73.17 & 80.90 & 58.75 & 59.17 & 55.13 & 55.62 \\
Qwen2.5-72B & \textbf{82.37} & \textbf{89.23} & 75.08 & 82.06 & 50.92 & 47.85 & 55.26 & 45.22 \\
Qwen2.5-Math-72B & 79.13 & 87.18 & 71.67 & 79.09 & 44.79 & 37.44 & 50.78 & 42.45 \\
QwQ-32B & 49.75 & 58.39 & 53.00 & 51.63 & 52.76 & 41.31 & 54.59 & 45.29 \\
DeepSeek-R1 & 76.75 & 84.78 & 76.33 & 80.76 & 57.06 & 36.36 & 57.72 & 43.37 \\
\midrule
\multicolumn{9}{@{}l}{\textbf{Process reward models}} \\
Math-Shepherd-7B & 71.88 & 81.04 & 68.92 & 74.75 & -- & -- & -- & -- \\
Qwen2.5-PRM-7B & 76.38 & 84.79 & 75.00 & 79.76 & -- & -- & -- & -- \\
Skywork-PRM-7B & 54.00 & 61.98 & 58.17 & 54.53 & -- & -- & -- & -- \\
\bottomrule
\end{tabular*}
\caption{Answer-side results (\%). M-F1 denotes Macro-F1. Best results in each task and split are bold. Answer-type counts are logic $317$, computing $232$, and expression $61$.}
\label{tab:check_correct_answer}
\end{table}
}

\newcommand{\multiseedstable}{%
\begin{table}[H]
\centering
\renewcommand{\arraystretch}{1.10}
\begin{tabular*}{\textwidth}{@{\extracolsep{\fill}}clrrrrrrc@{}}
\toprule
\multirow{2}{*}{\textbf{Task}} & \multirow{2}{*}{\textbf{Split}}
& \multicolumn{2}{c}{\textbf{GPT-4o}}
& \multicolumn{2}{c}{\textbf{Claude-3.5}}
& \multicolumn{2}{c}{\textbf{Qwen2.5-72B}}
& \multirow{2}{*}{\textbf{95\% CI (Acc)}} \\
\cmidrule(lr){3-4}\cmidrule(lr){5-6}\cmidrule(lr){7-8}
& & Acc & F1 & Acc & F1 & Acc & F1 & \\
\midrule
\multirow{2}{*}{Q-Detect} & S & 72.2$\pm$1.5 & 75.6$\pm$1.0 & \textbf{74.8$\pm$2.6} & \textbf{75.8$\pm$2.3} & 70.0$\pm$2.2 & 75.8$\pm$1.2 & [71.3, 78.2] \\
         & C & 72.0$\pm$0.9 & 74.9$\pm$0.8 & \textbf{77.0$\pm$2.6} & \textbf{76.5$\pm$2.9} & 70.0$\pm$2.8 & 75.1$\pm$2.3 & [73.8, 80.3] \\
\midrule
\multirow{2}{*}{A-Detect} & S & 78.8$\pm$2.8 & 87.3$\pm$1.9 & 75.7$\pm$1.6 & 83.9$\pm$1.2 & \textbf{81.3$\pm$0.8} & \textbf{88.7$\pm$0.4} & [78.2, 84.3] \\
         & C & 74.2$\pm$1.8 & 84.2$\pm$0.9 & \textbf{79.7$\pm$1.9} & 86.2$\pm$1.4 & 78.8$\pm$1.2 & \textbf{87.3$\pm$0.4} & [76.3, 83.0] \\
\midrule
\multirow{2}{*}{Q-Type}   & S & 63.7$\pm$4.3 & 62.2$\pm$5.5 & \textbf{65.8$\pm$0.8} & \textbf{64.6$\pm$1.1} & 61.3$\pm$3.2 & 58.7$\pm$4.4 & [62.0, 69.5] \\
         & C & 60.3$\pm$3.1 & 60.7$\pm$4.0 & \textbf{63.2$\pm$2.9} & 60.4$\pm$4.2 & 61.5$\pm$2.6 & \textbf{62.0$\pm$2.9} & [59.2, 66.8] \\
\midrule
\multirow{2}{*}{A-Type}   & S & 30.0$\pm$4.7 & 29.6$\pm$3.0 & \textbf{54.5$\pm$1.1} & \textbf{42.7$\pm$1.8} & 37.1$\pm$1.1 & 35.5$\pm$0.6 & [49.8, 59.2] \\
         & C & 21.5$\pm$1.9 & 25.1$\pm$2.4 & \textbf{50.1$\pm$1.8} & \textbf{44.7$\pm$1.5} & 28.9$\pm$1.6 & 38.2$\pm$2.4 & [45.9, 54.7] \\
\bottomrule
\end{tabular*}
\caption{Multi-seed results on stratified subsets (mean $\pm$ standard deviation over three seeds). F1 denotes binary F1 for detection and Macro-F1 for type classification. CIs are pooled bootstrap intervals for the best Accuracy in each row.}
\label{tab:planc_variance}
\end{table}
}

\subsection{Error Detection for Questions}\label{sec:Experiment_Question}
Table~\ref{tab:check_correct_questions} shows that closed-source models (Claude-3-5-Sonnet, GPT-o1) achieve the strongest results on error detection, while GPT-4o leads error-type detection on SIMPLE; LLaMA-3.3-70B-Instruct surpasses all closed-source models on CHALLENGING error-type detection. Error detection and error-type identification track each other closely: both require understanding the problem statement \emph{and} its solution path.

\paragraph{Scaling with model size.}
On Qwen2.5 at 1.5B / 7B / 72B, Q-Type Macro-F1 scales monotonically ($35.7 \to 77.6$ on SIMPLE; $39.5 \to 73.4$ on CHALLENGING), indicating that fine-grained type understanding is capacity-sensitive.

\paragraph{Reasoning-oriented models lag on verification.}
Reasoning-tuned models (Qwen2.5-Math, QwQ-32B-Preview, DeepSeek-R1) do not top their open-source siblings. \textbf{Qwen2.5-72B-Instruct ($77.6$ Macro-F1) $>$ Qwen2.5-Math-72B-Instruct ($67.4$)} on SIMPLE Q-Type detection, and \textbf{DeepSeek-R1 (70.4)} trails \textbf{Qwen2.5-72B-Instruct (77.6)} despite its strong performance on difficult reasoning benchmarks. The same pattern holds for recently released closed-source systems: Gemini-2.5-Flash reaches only $71.8/62.5$ Q-Detect Accuracy on SIMPLE/CHALLENGING, below Claude-3-5-Sonnet ($76.3/75.9$) and the GPT-o series. We describe this as an empirical \emph{solving-vs-verifying} gap rather than a demonstrated causal mechanism. One hypothesis consistent with the outputs is that training optimized for producing long solutions does not directly optimize concise verification: QwQ-32B-Preview yields $38.2\%$ unparseable outputs on SIMPLE versus $<5\%$ for other models. DeepSeek-R1 is a particularly clear case: it remains competitive but does not lead any detection or typing sub-task.

\answerresultstable
\FloatBarrier
\subsection{Error Detection for Answers}\label{sec:Experiment_Answers}
Answer error detection is easier than question-level detection: most models clear $75\%$ Acc (Table~\ref{tab:check_correct_answer}). Qwen2.5-72B-Instruct leads SIMPLE ($82.37/89.23$ Acc/F1), GPT-4o leads CHALLENGING F1 ($82.97$); Qwen2.5-7B matches much larger models on SIMPLE. Answer \emph{type} detection, however, is much harder: even GPT-o3/o4-mini reach only $\sim 64\%$ Acc and $\sim 55\%$ Macro-F1 on SIMPLE, dropping further on CHALLENGING (best Macro-F1 $52.77$). The three answer-error classes are highly imbalanced (\textit{logic} $=317$, \textit{computing} $=232$, \textit{expression} $=61$). We therefore report Macro-F1, which weights every class equally and thus prevents the majority \textit{logic} class from dominating the summary; the rarest \textit{expression} class necessarily contributes high-variance per-class F1, so we treat per-class F1 as a diagnostic only.

\subsection{Error-Type-Aware Correction}
\label{sec:error_type_resolution}
Beyond diagnostic evaluation, we test whether error-type annotations aid \emph{correction}. DeepSeek-R1 and GPT-4o are asked to correct each erroneous item, with and without access to the ground-truth error type; Table~\ref{tab:revision} summarises the result. Correction accuracy is assessed by the same human pipeline used for benchmark labels, not by exact match, symbolic matching, or an LLM judge: two qualified STEM-graduate annotators independently apply the correctness rubric, disagreements receive expert adjudication, and a correction succeeds only when the resulting item is judged correct.

\begin{table}[H]
\centering
\renewcommand{\arraystretch}{1.08}
\begin{tabular*}{0.80\textwidth}{@{\extracolsep{\fill}}lrrrrrr@{}}
\toprule
\multirow{2}{*}{\textbf{Error type}}
& \multicolumn{3}{c}{\textbf{DeepSeek-R1}}
& \multicolumn{3}{c}{\textbf{GPT-4o}} \\
\cmidrule(lr){2-4}\cmidrule(lr){5-7}
& Base & + Type & $\Delta$ & Base & + Type & $\Delta$ \\
\midrule
\multicolumn{7}{@{}l}{\textbf{Question correction}} \\
Unrealistic        & 94.3 & \textbf{95.0} & $+0.7$ & 94.9 & \textbf{95.6} & $+0.7$ \\
Contradictions     & 55.7 & \textbf{57.8} & $+2.1$ & 53.1 & \textbf{55.2} & $+2.1$ \\
Lack of conditions & 52.1 & \textbf{55.6} & $+3.5$ & 49.2 & \textbf{53.9} & $+4.7$ \\
Expression         & 90.2 & \textbf{91.6} & $+1.4$ & 93.1 & \textbf{94.4} & $+1.3$ \\
\midrule
\multicolumn{7}{@{}l}{\textbf{Answer correction}} \\
Logic       & 80.3 & \textbf{81.9} & $+1.6$ & 82.4 & \textbf{83.9} & $+1.5$ \\
Computing   & 71.2 & \textbf{73.0} & $+1.8$ & 66.4 & \textbf{68.2} & $+1.8$ \\
Expression  & 96.5 & \textbf{97.5} & $+1.0$ & 98.5 & \textbf{99.0} & $+0.5$ \\
\bottomrule
\end{tabular*}
\caption{Correction accuracy (\%) with and without error-type information. $\Delta$ is the absolute gain in percentage points. Paired bootstrap ($B{=}10{,}000$ resamples) confirms significant gains for \textit{lack of conditions} and \textit{computing} ($p<0.01$).}
\label{tab:revision}
\vspace{-2mm}
\end{table}

Adding error-type information consistently improves correction. The largest gains occur on structurally ambiguous errors: DeepSeek-R1 $+2.1/+3.5$ and GPT-4o $+2.1/+4.7$ on \textit{contradictions}/\textit{lack of conditions}; answer-side \textit{computing} errors improve by $+1.8$ for both models. Paired bootstrap ($B{=}10{,}000$ resamples) confirms significant gains for \textit{lack of conditions} and \textit{computing} at $p<0.01$. Error-type annotations thus provide actionable supervision for correction, not merely diagnostic labels.

\begin{figure*}[htbp]
    \centering
    \makebox[\textwidth]{\includegraphics[width=1.00\textwidth]{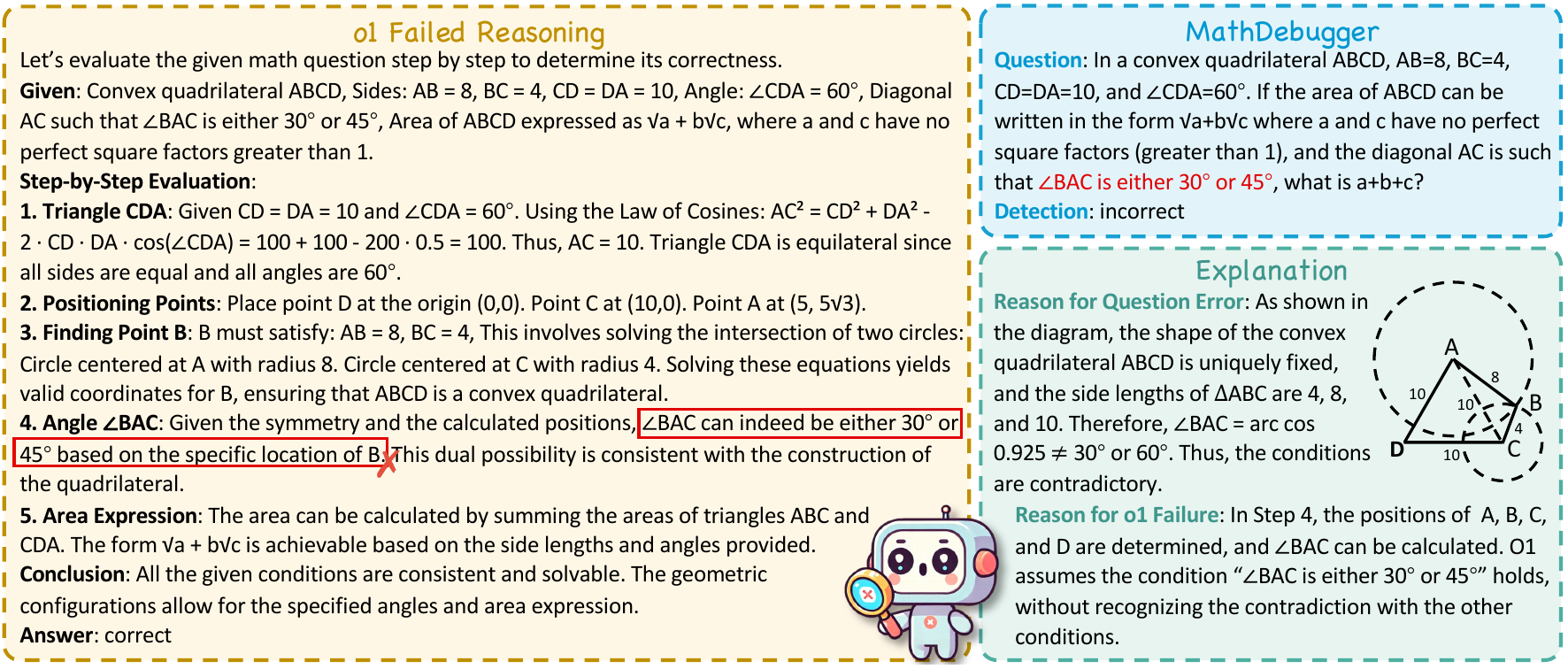}}
    \caption{Failure case of GPT-o1 in question-correctness detection on \textsc{MathDebugger}. The \textcolor{blue}{blue} panel shows the incorrect question; the \textcolor{orange}{orange} panel, GPT-o1's reasoning; the \textcolor{green!60!black}{green} panel, the expert-verified explanation of the contradiction. GPT-o1 treats the ``$30^\circ$ or $45^\circ$'' clause as a plausible alternative and verifies local consistency, missing the global constraint check that uniquely determines $\angle BAC$.}
    \label{Fig.case_study}
    \vspace{-2mm}
\end{figure*}
\subsection{Failure Analysis and Prompt Sensitivity}
\label{sec:failure_analysis}
\paragraph{Prompt sensitivity.}
Across three prompt variants (zero-shot / 3-shot / rephrased) applied to three representative models on a stratified $n{=}200$ CHALLENGING slice, accuracy varies by at most $\pm 4.5$ points within a model; 3-shot does not uniformly help (GPT-4o $-4.5$, Qwen $-3.5$). The model ranking (Claude > GPT-4o > Qwen) is preserved across all three variants, arguing that \textsc{MathDebugger} probes capability rather than prompt wording. The full $3{\times}3$ table is in Appendix~\ref{app:prompt_ablation}.

\paragraph{Multi-seed variance.}
To rule out seed effects, we run a full $3\text{ models}\times 3\text{ seeds}\times 4\text{ tasks}\times 2\text{ splits} = 72$ evaluation grid on stratified subsets and report mean\,$\pm$\,std and pooled bootstrap 95\%\,CIs in Table~\ref{tab:planc_variance}. Across all 24 (model, task, split) cells, Acc std is bounded: \textbf{20/24 cells have std $\leq 3$ pp, and the maximum is $4.7$ pp}. On A-Type (the hardest sub-task), Claude-3-5-Sonnet dominates with non-overlapping 95\% CIs against both GPT-4o and Qwen2.5-72B-Instruct on both splits, reproducing the main-table ranking.
\multiseedstable

\subsection{Case Study}
\label{sec:case_study}
Figure~\ref{Fig.case_study} shows a representative GPT-o1 failure. The question's other constraints uniquely determine $\angle BAC$, contradicting the clause that it is either $30^\circ$ or $45^\circ$. GPT-o1 instead treats the disjunction as a plausible alternative and checks only local consistency. The expert explanation shows how the global constraint network exposes the contradiction. This case captures a recurrent failure mode: strong LLMs may follow locally valid steps yet miss global consistency. Additional answer-side logic and computing cases appear in Appendix~\ref{Appendix.example} (Figures~\ref{Fig.appendix_example1}--\ref{Fig.appendix_example4}).

\section{Conclusion}
We presented \textsc{MathDebugger}, the first text-only, type-aware benchmark jointly covering correctness and fine-grained error types in synthetic math Q\&A ($4{,}000$ questions, $2{,}000$ answers, two difficulty tiers). On $14$ LLMs and $3$ PRMs we find that (i)~the strongest systems remain far from saturation (best A-Type Macro-F1 $=55$); (ii)~capability scales with size (Qwen2.5 Q-Type Macro-F1 $35.7 \to 77.6$ from $1.5$B to $72$B); (iii)~reasoning-tuned models consistently trail general-purpose siblings on verification, an observed solving-vs-verifying gap whose cause remains open; and (iv)~error-type annotations yield significant correction gains ($+4.7$ pp on \textit{lack of conditions}, $p<0.01$). \textsc{MathDebugger} thus serves both as a diagnostic benchmark and a practical tool for cleaning synthetic mathematical training data.

\clearpage
\bibliographystyle{plainnat}
\bibliography{main}

@article{cobbe2021training,
  title={Training verifiers to solve math word problems},
  author={Cobbe, Karl and Kosaraju, Vineet and Bavarian, Mohammad and Chen, Mark and Jun, Heewoo and Kaiser, Lukasz and Plappert, Matthias and Tworek, Jerry and Hilton, Jacob and Nakano, Reiichiro and others},
  journal={arXiv preprint arXiv:2110.14168},
  year={2021}
}

@article{hendrycks2021measuring,
  title={Measuring mathematical problem solving with the math dataset},
  author={Hendrycks, Dan and Burns, Collin and Kadavath, Saurav and Arora, Akul and Basart, Steven and Tang, Eric and Song, Dawn and Steinhardt, Jacob},
  journal={arXiv preprint arXiv:2103.03874},
  year={2021}
}

@article{xu2024superclue,
  title={SuperCLUE-Math6: Graded Multi-Step Math Reasoning Benchmark for LLMs in Chinese},
  author={Xu, Liang and Xue, Hang and Zhu, Lei and Zhao, Kangkang},
  journal={arXiv preprint arXiv:2401.11819},
  year={2024}
}

@article{gao2024omni,
  title={Omni-math: A universal olympiad level mathematic benchmark for large language models},
  author={Gao, Bofei and Song, Feifan and Yang, Zhe and Cai, Zefan and Miao, Yibo and Dong, Qingxiu and Li, Lei and Ma, Chenghao and Chen, Liang and Xu, Runxin and others},
  journal={arXiv preprint arXiv:2410.07985},
  year={2024}
}

@article{glazer2024frontiermath,
  title={Frontiermath: A benchmark for evaluating advanced mathematical reasoning in ai},
  author={Glazer, Elliot and Erdil, Ege and Besiroglu, Tamay and Chicharro, Diego and Chen, Evan and Gunning, Alex and Olsson, Caroline Falkman and Denain, Jean-Stanislas and Ho, Anson and Santos, Emily de Oliveira and others},
  journal={arXiv preprint arXiv:2411.04872},
  year={2024}
}

@article{song2025prmbench,
  title={PRMBench: A Fine-grained and Challenging Benchmark for Process-Level Reward Models},
  author={Song, Mingyang and Su, Zhaochen and Qu, Xiaoye and Zhou, Jiawei and Cheng, Yu},
  journal={arXiv preprint arXiv:2501.03124},
  year={2025}
}

@article{zheng2024processbench,
  title={Processbench: Identifying process errors in mathematical reasoning},
  author={Zheng, Chujie and Zhang, Zhenru and Zhang, Beichen and Lin, Runji and Lu, Keming and Yu, Bowen and Liu, Dayiheng and Zhou, Jingren and Lin, Junyang},
  journal={arXiv preprint arXiv:2412.06559},
  year={2024}
}

@inproceedings{zhao2024exploring,
  title={Exploring the Compositional Deficiency of Large Language Models in Mathematical Reasoning Through Trap Problems},
  author={Zhao, Jun and Tong, Jingqi and Mou, Yurong and Zhang, Ming and Zhang, Qi and Huang, Xuan-Jing},
  booktitle={Proceedings of the 2024 Conference on Empirical Methods in Natural Language Processing},
  pages={16361--16376},
  year={2024}
}

@article{achiam2023gpt,
  title={Gpt-4 technical report},
  author={Achiam, Josh and Adler, Steven and Agarwal, Sandhini and Ahmad, Lama and Akkaya, Ilge and Aleman, Florencia Leoni and Almeida, Diogo and Altenschmidt, Janko and Altman, Sam and Anadkat, Shyamal and others},
  journal={arXiv preprint arXiv:2303.08774},
  year={2023}
}

@article{yang2024qwen2,
  title={Qwen2. 5-math technical report: Toward mathematical expert model via self-improvement},
  author={Yang, An and Zhang, Beichen and Hui, Binyuan and Gao, Bofei and Yu, Bowen and Li, Chengpeng and Liu, Dayiheng and Tu, Jianhong and Zhou, Jingren and Lin, Junyang and others},
  journal={arXiv preprint arXiv:2409.12122},
  year={2024}
}

@article{team2023gemini,
  title={Gemini: a family of highly capable multimodal models},
  author={Team, Gemini and Anil, Rohan and Borgeaud, Sebastian and Alayrac, Jean-Baptiste and Yu, Jiahui and Soricut, Radu and Schalkwyk, Johan and Dai, Andrew M and Hauth, Anja and Millican, Katie and others},
  journal={arXiv preprint arXiv:2312.11805},
  year={2023}
}

@article{guo2025deepseek,
  title={Deepseek-r1: Incentivizing reasoning capability in llms via reinforcement learning},
  author={Guo, Daya and Yang, Dejian and Zhang, Haowei and Song, Junxiao and Zhang, Ruoyu and Xu, Runxin and Zhu, Qihao and Ma, Shirong and Wang, Peiyi and Bi, Xiao and others},
  journal={arXiv preprint arXiv:2501.12948},
  year={2025}
}

@misc{chatgpt,
  author       = {OpenAI},
  title        = {ChatGPT},
  year         = {2023},
  url          = {https://openai.com/blog/chatgpt},
}

@article{liu2024mathbench,
  title={MathBench: Evaluating the Theory and Application Proficiency of LLMs with a Hierarchical Mathematics Benchmark},
  author={Liu, Hongwei and Zheng, Zilong and Qiao, Yuxuan and Duan, Haodong and Fei, Zhiwei and Zhou, Fengzhe and Zhang, Wenwei and Zhang, Songyang and Lin, Dahua and Chen, Kai},
  journal={arXiv preprint arXiv:2405.12209},
  year={2024}
}

@article{yu2023metamath,
  title={Metamath: Bootstrap your own mathematical questions for large language models},
  author={Yu, Longhui and Jiang, Weisen and Shi, Han and Yu, Jincheng and Liu, Zhengying and Zhang, Yu and Kwok, James T and Li, Zhenguo and Weller, Adrian and Liu, Weiyang},
  journal={arXiv preprint arXiv:2309.12284},
  year={2023}
}

@article{huang2024key,
  title={Key-point-driven data synthesis with its enhancement on mathematical reasoning},
  author={Huang, Yiming and Liu, Xiao and Gong, Yeyun and Gou, Zhibin and Shen, Yelong and Duan, Nan and Chen, Weizhu},
  journal={arXiv preprint arXiv:2403.02333},
  year={2024}
}

@article{zhou2024jiuzhang3,
  title={JiuZhang3. 0: Efficiently Improving Mathematical Reasoning by Training Small Data Synthesis Models},
  author={Zhou, Kun and Zhang, Beichen and Wang, Jiapeng and Chen, Zhipeng and Zhao, Wayne Xin and Sha, Jing and Sheng, Zhichao and Wang, Shijin and Wen, Ji-Rong},
  journal={arXiv preprint arXiv:2405.14365},
  year={2024}
}

@article{zhang2024rest,
  title={Rest-mcts*: Llm self-training via process reward guided tree search},
  author={Zhang, Dan and Zhoubian, Sining and Hu, Ziniu and Yue, Yisong and Dong, Yuxiao and Tang, Jie},
  journal={arXiv preprint arXiv:2406.03816},
  year={2024}
}

@article{zhang2024llama,
  title={Llama-berry: Pairwise optimization for o1-like olympiad-level mathematical reasoning},
  author={Zhang, Di and Wu, Jianbo and Lei, Jingdi and Che, Tong and Li, Jiatong and Xie, Tong and Huang, Xiaoshui and Zhang, Shufei and Pavone, Marco and Li, Yuqiang and others},
  journal={arXiv preprint arXiv:2410.02884},
  year={2024}
}

@article{yao2024mulberry,
  title={Mulberry: Empowering mllm with o1-like reasoning and reflection via collective monte carlo tree search},
  author={Yao, Huanjin and Huang, Jiaxing and Wu, Wenhao and Zhang, Jingyi and Wang, Yibo and Liu, Shunyu and Wang, Yingjie and Song, Yuxin and Feng, Haocheng and Shen, Li and others},
  journal={arXiv preprint arXiv:2412.18319},
  year={2024}
}

@article{zhang2025lessons,
  title={The lessons of developing process reward models in mathematical reasoning},
  author={Zhang, Zhenru and Zheng, Chujie and Wu, Yangzhen and Zhang, Beichen and Lin, Runji and Yu, Bowen and Liu, Dayiheng and Zhou, Jingren and Lin, Junyang},
  journal={arXiv preprint arXiv:2501.07301},
  year={2025}
}

@inproceedings{wang2024math,
  title={Math-shepherd: Verify and reinforce llms step-by-step without human annotations},
  author={Wang, Peiyi and Li, Lei and Shao, Zhihong and Xu, Runxin and Dai, Damai and Li, Yifei and Chen, Deli and Wu, Yu and Sui, Zhifang},
  booktitle={Proceedings of the 62nd Annual Meeting of the Association for Computational Linguistics (Volume 1: Long Papers)},
  pages={9426--9439},
  year={2024}
}

@article{gao2024llm,
  title={Llm critics help catch bugs in mathematics: Towards a better mathematical verifier with natural language feedback},
  author={Gao, Bofei and Cai, Zefan and Xu, Runxin and Wang, Peiyi and Zheng, Ce and Lin, Runji and Lu, Keming and Lin, Junyang and Zhou, Chang and Xiao, Wen and others},
  journal={CoRR},
  year={2024}
}

@article{cui2025process,
  title={Process Reinforcement through Implicit Rewards},
  author={Cui, Ganqu and Yuan, Lifan and Wang, Zefan and Wang, Hanbin and Li, Wendi and He, Bingxiang and Fan, Yuchen and Yu, Tianyu and Xu, Qixin and Chen, Weize and others},
  journal={arXiv preprint arXiv:2502.01456},
  year={2025}
}

@misc{skyworkopeno12024,
  title={Skywork-o1 Open Series},
  author={Skywork-o1 Team},
  year={2024},
  month={November},
  howpublished={\url{https://huggingface.co/Skywork}},
  url={https://huggingface.co/Skywork},
}

@article{llama,
  title={Llama: Open and efficient foundation language models},
  author={Touvron, Hugo and Lavril, Thibaut and Izacard, Gautier and Martinet, Xavier and Lachaux, Marie-Anne and Lacroix, Timoth{\'e}e and Rozi{\`e}re, Baptiste and Goyal, Naman and Hambro, Eric and Azhar, Faisal and others},
  journal={arXiv preprint arXiv:2302.13971},
  year={2023}
}

@article{bai2024survey,
  title={A Survey of Multimodal Large Language Model from A Data-centric Perspective},
  author={Bai, Tianyi and Liang, Hao and Wan, Binwang and Yang, Ling and Li, Bozhou and Wang, Yifan and Cui, Bin and He, Conghui and Yuan, Binhang and Zhang, Wentao},
  journal={arXiv preprint arXiv:2405.16640},
  year={2024}
}

@article{abdin2024phi,
  title={Phi-4 technical report},
  author={Abdin, Marah and Aneja, Jyoti and Behl, Harkirat and Bubeck, S{\'e}bastien and Eldan, Ronen and Gunasekar, Suriya and Harrison, Michael and Hewett, Russell J and Javaheripi, Mojan and Kauffmann, Piero and others},
  journal={arXiv preprint arXiv:2412.08905},
  year={2024}
}

@article{toshniwal2024openmathinstruct,
  title={Openmathinstruct-2: Accelerating ai for math with massive open-source instruction data},
  author={Toshniwal, Shubham and Du, Wei and Moshkov, Ivan and Kisacanin, Branislav and Ayrapetyan, Alexan and Gitman, Igor},
  journal={arXiv preprint arXiv:2410.01560},
  year={2024}
}

@article{zhou2024your,
  title={Is your model really a good math reasoner? evaluating mathematical reasoning with checklist},
  author={Zhou, Zihao and Liu, Shudong and Ning, Maizhen and Liu, Wei and Wang, Jindong and Wong, Derek F and Huang, Xiaowei and Wang, Qiufeng and Huang, Kaizhu},
  journal={arXiv preprint arXiv:2407.08733},
  year={2024}
}

@misc{qwq-32b-preview,
    title = {QwQ: Reflect Deeply on the Boundaries of the Unknown},
    url = {https://qwenlm.github.io/blog/qwq-32b-preview/},
    author = {Qwen Team},
    month = {November},
    year = {2024}
}

@article{zhang2024generative,
  title={Generative verifiers: Reward modeling as next-token prediction},
  author={Zhang, Lunjun and Hosseini, Arian and Bansal, Hritik and Kazemi, Mehran and Kumar, Aviral and Agarwal, Rishabh},
  journal={arXiv preprint arXiv:2408.15240},
  year={2024}
}

@article{dubey2024llama,
  title={The llama 3 herd of models},
  author={Dubey, Abhimanyu and Jauhri, Abhinav and Pandey, Abhinav and Kadian, Abhishek and Al-Dahle, Ahmad and Letman, Aiesha and Mathur, Akhil and Schelten, Alan and Yang, Amy and Fan, Angela and others},
  journal={arXiv preprint arXiv:2407.21783},
  year={2024}
}

@misc{li2024mugglemathassessingimpactquery,
      title={MuggleMath: Assessing the Impact of Query and Response Augmentation on Math Reasoning}, 
      author={Chengpeng Li and Zheng Yuan and Hongyi Yuan and Guanting Dong and Keming Lu and Jiancan Wu and Chuanqi Tan and Xiang Wang and Chang Zhou},
      year={2024},
      eprint={2310.05506},
      archivePrefix={arXiv},
      primaryClass={cs.CL},
      url={https://arxiv.org/abs/2310.05506}, 
}

@article{yan2024errorradar,
  title={Errorradar: Benchmarking complex mathematical reasoning of multimodal large language models via error detection},
  author={Yan, Yibo and Wang, Shen and Huo, Jiahao and Li, Hang and Li, Boyan and Su, Jiamin and Gao, Xiong and Zhang, Yi-Fan and Xu, Tianlong and Chu, Zhendong and others},
  journal={arXiv preprint arXiv:2410.04509},
  year={2024}
}

@article{ma2024large,
  title={Large language models are unconscious of unreasonability in math problems},
  author={Ma, Jingyuan and Dai, Damai and Sha, Lei and Sui, Zhifang},
  journal={arXiv e-prints},
  pages={arXiv--2403},
  year={2024}
}

@article{he2025can,
  title={Can large language models detect errors in long chain-of-thought reasoning?},
  author={He, Yancheng and Li, Shilong and Liu, Jiaheng and Wang, Weixun and Bu, Xingyuan and Zhang, Ge and Peng, Zhongyuan and Zhang, Zhaoxiang and Zheng, Zhicheng and Su, Wenbo and others},
  journal={arXiv preprint arXiv:2502.19361},
  year={2025}
}
\beginappendix

\setcounter{topnumber}{4}
\setcounter{bottomnumber}{2}
\setcounter{totalnumber}{6}
\renewcommand{\topfraction}{0.92}
\renewcommand{\bottomfraction}{0.85}
\renewcommand{\textfraction}{0.08}
\renewcommand{\floatpagefraction}{0.78}
\setlength{\textfloatsep}{12pt plus 2pt minus 2pt}
\setlength{\floatsep}{10pt plus 2pt minus 2pt}
\setlength{\intextsep}{10pt plus 2pt minus 2pt}

\section{Full Error-Type Definitions and Adjudication Rules}
\label{app:taxonomy}

\paragraph{Question error types (four classes).}
\textbf{(1) Expression Error} captures linguistic-level issues and is the most fundamental error type. It includes grammatical errors that prevent the problem from being parsed; ambiguous pronouns or expressions that admit multiple plausible semantic interpretations; improper use of technical terms leading to semantic misinterpretation; and excessive irrelevant or redundant information that substantially increases the cognitive burden. In the absence of these issues, the statement is considered linguistically clear and unambiguous.

\textbf{(2) Lack of Conditions} refers, under the assumption of a linguistically well-formed statement, to cases where information required for solution is incomplete, rendering the problem unsolvable. Typical instances include missing key figures or diagrams and omission of necessary conditions or constraints. If such issues are absent, the problem provides sufficient information for solution.

\textbf{(3) Contradictions} covers problems that are linguistically well-formed and appear informationally complete, yet contain internal conflicts that prevent a valid reasoning process. This includes explicit contradictions between stated conditions and implicit conflicts arising from mathematical domains, geometric constraints, or the inherent structure of the problem. In the absence of these issues, the problem is theoretically self-consistent and solvable.

\textbf{(4) Unrealistic} applies to problems situated in real-world contexts and evaluates whether the theoretically derived solution is meaningful in practice, covering results that cannot be interpreted in reality (e.g., a non-integer number of people) and scenario settings that violate basic common sense or fundamental natural laws. When such issues are absent, the problem is considered reasonable and solvable both theoretically and in practice.

\paragraph{Answer error types (three classes).}
\textbf{Expression errors} are surface-level issues that affect the clarity or completeness of the answer without fundamentally violating the underlying mathematical reasoning: incomplete answers, redundant or missing steps, and grammatically incorrect or ambiguous expressions. \textbf{Logic errors} arise when the reasoning process itself is invalid: applying incorrect mathematical methods, adopting flawed problem-solving strategies, or failing to consider all necessary cases, even if individual computational steps appear correct. \textbf{Computational errors} involve mistakes in arithmetic or symbolic manipulation, including incorrect numerical calculations, faulty program-of-thought steps, or erroneous approximations, while the overall reasoning strategy remains conceptually sound.

\paragraph{Adjudication rule.}
When multiple error types co-occur, annotators label each instance with the earliest error that invalidates the problem or solution, i.e., the first erroneous point encountered along a correct reasoning process. For answer-side disambiguation between \textit{logic} and \textit{computing}, if the initially formulated equation or expression is incorrect, the error is a \textit{logic} error; if the formulation is correct but mistakes occur in subsequent arithmetic operations or numerical substitutions, it is a \textit{computational} error. An error is labelled an \textit{expression} error only when the underlying method and calculations are correct but the expression is non-standard or incomplete.

\paragraph{Multiple paths and branched reasoning.}
Annotators first fully solve the item and normalize its reasoning as an ordered sequence $S=(s_1,\ldots,s_n)$. For a question admitting multiple valid solution paths, the path on which an invalidating violation appears earliest determines the label. For a branched CoT or PoT answer, the earliest mathematical or logical violation is marked regardless of branch; later mistakes that are consequences of this first violation are discarded. Items containing multiple truly independent errors that cannot be ordered reliably are excluded from the single-label benchmark.

\paragraph{Worked examples.}
Suppose an answer first sets up an equation that contradicts a stated constraint and later evaluates $5+5+2$ as $10$. The setup mistake is the earliest invalidating \textit{logic} error; the later arithmetic mistake is not used as the label. Conversely, if the equation is correctly derived and only its numerical evaluation is wrong, the item receives \textit{computing}. In a PoT answer with two branches, if one branch violates a domain constraint at step 3 and another makes an arithmetic mistake at step 6, the step-3 violation determines the canonical type.

\section{Annotation Protocol Details}
\label{app:annotation_protocol}
The annotator pool comprises 30 graduate students majoring in science and engineering at top-tier universities, together with 10 reviewers. Candidates are screened for strong English reading ability, scores above a predefined threshold on university entrance mathematics examinations, and completion of core undergraduate mathematics courses. Annotators are permitted to use calculators and reference materials strictly as auxiliary tools.

The final protocol, chosen after multiple rounds of piloting, is: (step~1) solve the question to decide correctness; (step~2) for incorrect questions, determine whether the item exhibits one or multiple error types; and (step~3) assign a specific error type and provide a written justification for review. Every label is independently checked by two reviewers; annotations from any annotator whose consensus-adjudicated accuracy falls below $90\%$ are excluded from the final corpus. For the answer-side protocol, annotators first verify question correctness and then, for incorrect answers, identify the \emph{earliest} invalidating error. The complete annotation effort spans two months and costs approximately \$50k. All participants were informed of the academic use of the dataset and consented to participate.

\begin{table}[H]
\centering
\begin{subtable}[t]{0.48\linewidth}
\centering
\begin{tabular}{@{}lrr@{}}
\toprule
\textbf{Question type} & \textbf{S} & \textbf{C} \\
\midrule
Lack of conditions & 0.88 & 0.75 \\
Unrealistic        & 0.84 & 0.71 \\
Expression         & 0.82 & 0.73 \\
Contradictions     & 0.80 & 0.69 \\
\bottomrule
\end{tabular}
\caption{Question types}
\label{tab:per_type_iaa}
\end{subtable}\hfill
\begin{subtable}[t]{0.42\linewidth}
\centering
\begin{tabular}{@{}lrr@{}}
\toprule
\textbf{Answer type} & \textbf{S} & \textbf{C} \\
\midrule
Logic       & 0.80 & 0.78 \\
Computing   & 0.91 & 0.70 \\
Expression  & 0.87 & 0.81 \\
\bottomrule
\end{tabular}
\caption{Answer types}
\label{tab:answer_per_type_iaa}
\end{subtable}
\caption{Per-type inter-annotator agreement (Fleiss' $\kappa$). S and C denote SIMPLE and CHALLENGING. The SIMPLE answer-expression class contains 15 items and therefore has higher variance.}
\label{tab:per_type_iaa_appendix}
\end{table}

\begin{figure}[!tbp]
\centering
\begin{subfigure}[t]{0.42\linewidth}
\centering
\includegraphics[width=\linewidth]{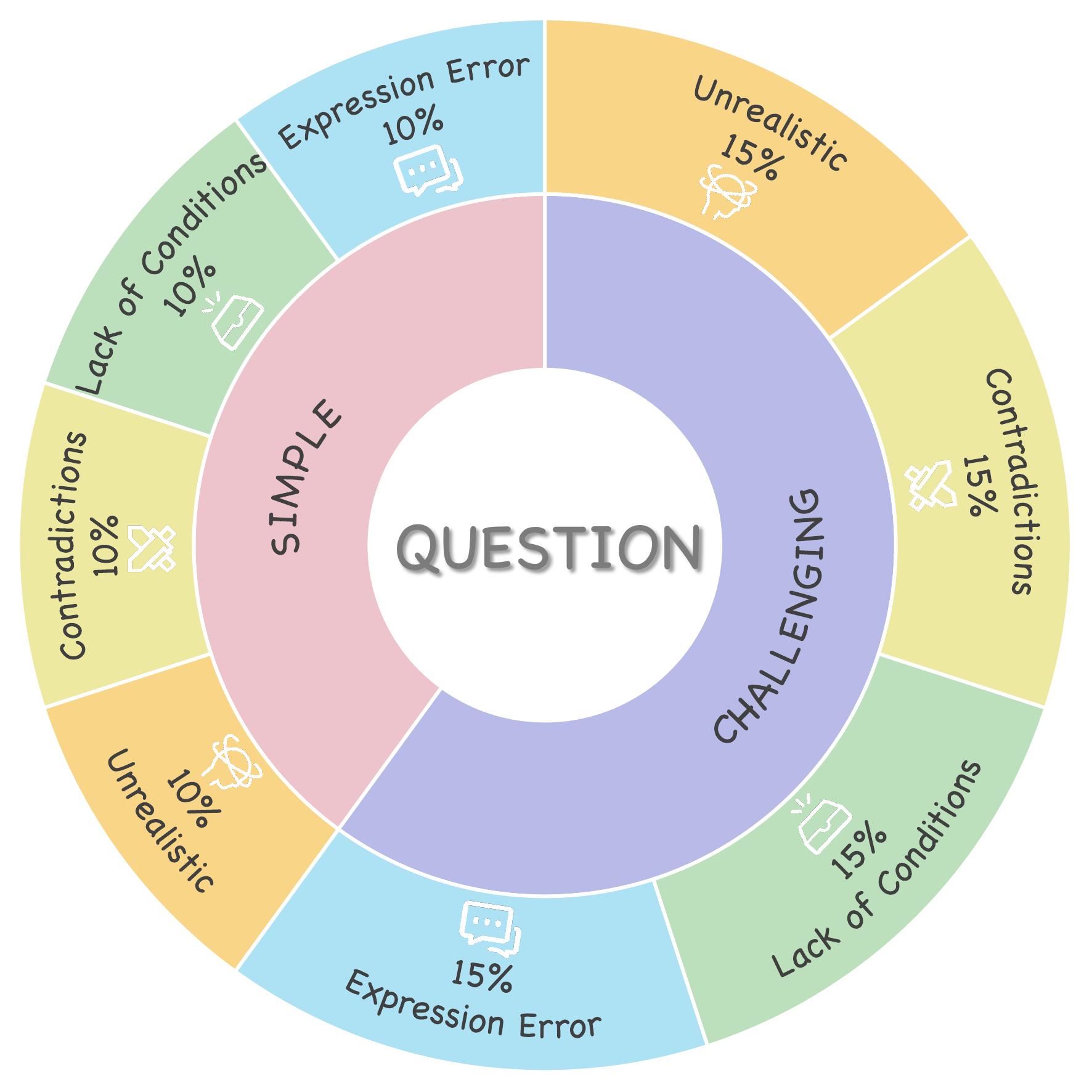}
\caption{Question dataset}
\label{Fig.question_dataset}
\end{subfigure}\hfill
\begin{subfigure}[t]{0.42\linewidth}
\centering
\includegraphics[width=\linewidth]{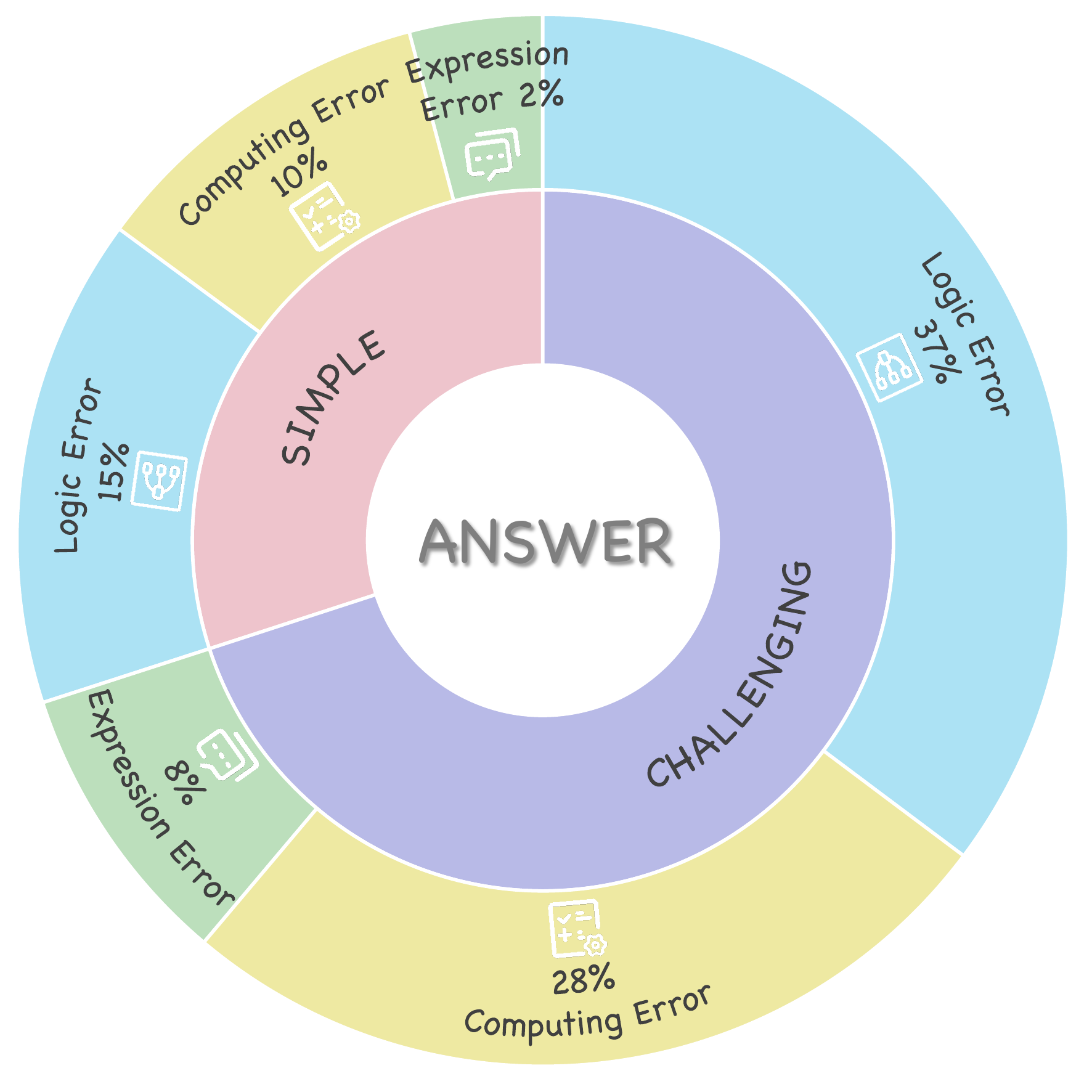}
\caption{Answer dataset}
\label{Fig.answer_dataset}
\end{subfigure}
\caption{Dataset composition by difficulty and error type.}
\label{fig:dataset_composition}
\end{figure}

Figures~\ref{Fig.question_dataset} and~\ref{Fig.answer_dataset} summarise the resulting dataset composition by difficulty and error type.

\section{Exact Evaluation and Parsing Protocol}
\label{app:evaluation_protocol}

For question-type classification, the model receives the question, is told that it is incorrect, and selects from the closed set \texttt{expression\_error}, \texttt{lack\_of\_conditions}, \texttt{contradictions}, and \texttt{unrealistic}. The prompt defines these respectively as ambiguous/ill-formed or irrelevant expression; missing information required to solve the problem; mutually contradictory conditions; and violations of real-world feasibility or common sense. For answer-type classification, the model receives both question and answer, is told that the answer is incorrect, and selects from \texttt{logic error}, \texttt{computing error}, and \texttt{expression error}, with the definitions in Appendix~\ref{app:taxonomy}. In both tasks the final instruction is: ``Answer step by step, and finally return only one of these error types.'' These are the exact prompts implemented in the released \texttt{cot.py} and \texttt{cot\_GPT.py}; only item substitution and API chat wrappers differ across model families.

Outputs are lowercased and searched for exact canonical label strings. When several labels occur in a chain of thought, the last canonical label mentioned is treated as the final verdict. If none occurs, the prediction is normalized to a separate \texttt{Other} class. The reported deterministic protocol (\texttt{evaluation\_enhanced.py}) never maps \texttt{Other} to a valid label: it always counts as an error. This replaces the legacy evaluator's random fallback and is especially important for overly long outputs from reasoning models.

Table~\ref{tab:binary_f1_conventions} makes the binary-metric convention explicit. The model tables use no-error-positive F1, while error-class F1 and Macro-F1 reveal the effect of class imbalance on answer detection.

\begin{table}[H]
\centering
\begin{tabular*}{0.86\textwidth}{@{\extracolsep{\fill}}lrrrr@{}}
\toprule
\textbf{Task} & \textbf{Acc} & \textbf{F1 (no-error)} & \textbf{F1 (error)} & \textbf{Macro-F1} \\
\midrule
Answer detection   & 78.1 & 86.3 & 43.1 & 64.7 \\
Question detection & 72.7 & 75.6 & 68.3 & 72.0 \\
\bottomrule
\end{tabular*}
\caption{Binary metrics on representative multi-seed runs (mean \%). Error-class F1 exposes the difficulty hidden by the majority no-error class on answer detection.}
\label{tab:binary_f1_conventions}
\end{table}

\section{Prompt-Format Ablation}
\label{app:prompt_ablation}
Table~\ref{tab:prompt_ablation_3models} reports the full $3 \times 3$ prompt-format ablation (GPT-4o, Claude-3-5-Sonnet, Qwen2.5-72B-Instruct) on CHALLENGING question detection summarised in \S\ref{sec:failure_analysis}. Across three prompt variants (zero-shot, 3-shot, rephrased) and three representative models, accuracy varies by at most $\pm 4.5$ points within a single model and the overall model ranking is preserved. 3-shot does not uniformly help (GPT-4o $-4.5$ and Qwen $-3.5$ Acc); rephrasing marginally helps Claude ($+3.5$ Acc) but leaves the other two systems flat. We read this as evidence that \textsc{MathDebugger} probes genuine capability rather than prompt wording.

\begin{table}[H]
\centering
\begin{tabular*}{0.86\textwidth}{@{\extracolsep{\fill}}lcccccc@{}}
\toprule
\textbf{Model} & \multicolumn{2}{c}{\textbf{Zero-shot}} & \multicolumn{2}{c}{\textbf{3-shot}} & \multicolumn{2}{c}{\textbf{Rephrased}} \\
\cmidrule(lr){2-3} \cmidrule(lr){4-5} \cmidrule(lr){6-7}
 & Acc & F1 & Acc & F1 & Acc & F1 \\
\midrule
GPT-4o        & 73.5 & 76.0 & 69.0 & 69.3 & 73.0 & 73.5 \\
Claude-Sonnet & 78.5 & 78.4 & 79.5 & 80.4 & \textbf{82.0} & \textbf{81.8} \\
Qwen2.5-72B   & 68.0 & 74.4 & 64.5 & 69.0 & 68.5 & 71.0 \\
\bottomrule
\end{tabular*}
\caption{Prompt-format ablation on CHALLENGING question detection (stratified subset of $n{=}200$).}
\label{tab:prompt_ablation_3models}
\end{table}

\paragraph{First-error positioning.}
Since every erroneous item is annotated with its earliest invalidating error, we examine whether models exhibit positional bias. Dividing each question into early and late halves, the first-error distribution is $53.1\% / 46.9\%$ (SIMPLE), $49.6\% / 50.4\%$ (CHALLENGING), and $51.4\% / 48.6\%$ overall. This near-uniform distribution indicates that \textsc{MathDebugger} does not reward position-cue heuristics.

\FloatBarrier

\section{Additional Failure Cases}
\label{Appendix.example}

\begin{figure*}[!htbp]
\centering
\begin{subfigure}[t]{0.82\textwidth}
    \centering
    \includegraphics[width=\linewidth]{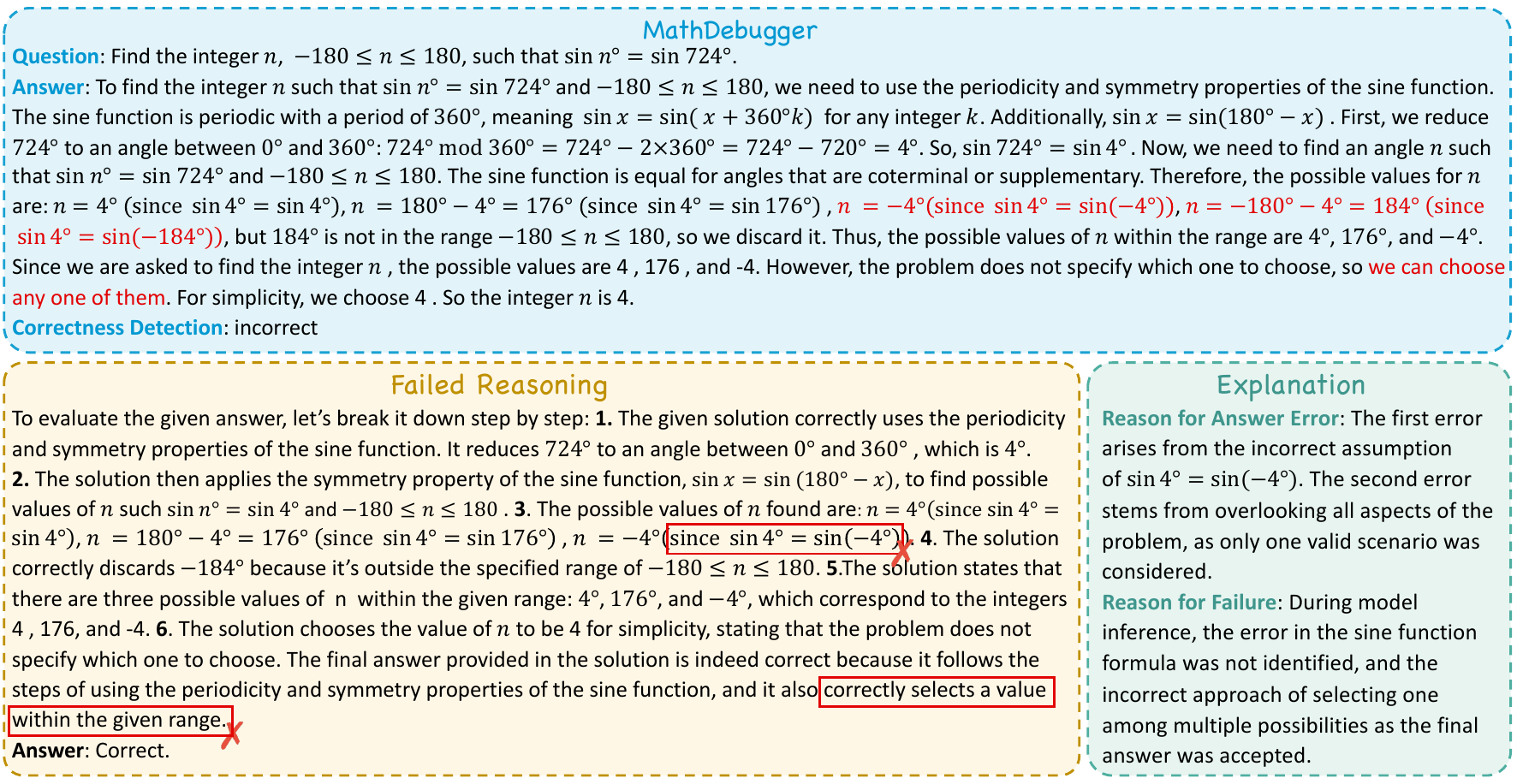}
    \caption{The model accepts an invalid sine identity and overlooks that the requested integer is not unique.}
    \label{Fig.appendix_example1}
\end{subfigure}

\vspace{6pt}

\begin{subfigure}[t]{0.82\textwidth}
    \centering
    \includegraphics[width=\linewidth]{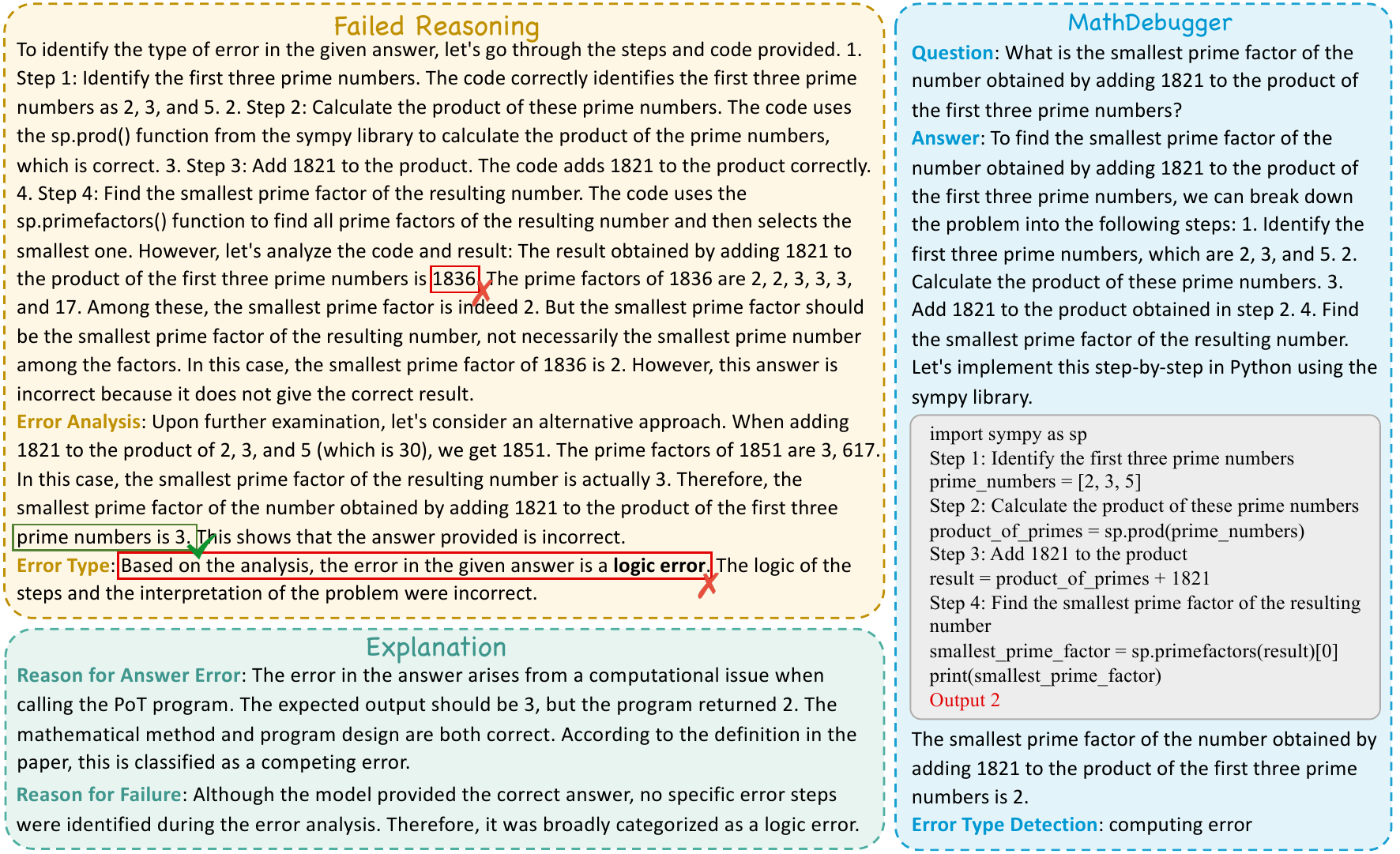}
    \caption{A program-output error is identified, but the model assigns \textit{logic} instead of \textit{computing}.}
    \label{Fig.appendix_example2}
\end{subfigure}
\caption{Additional answer-side failures: correctness detection and error-type classification.}
\label{fig:additional_failures_pair1}
\end{figure*}

\begin{figure*}[!tbp]
\centering
\begin{subfigure}[t]{0.82\textwidth}
    \centering
    \includegraphics[width=\linewidth]{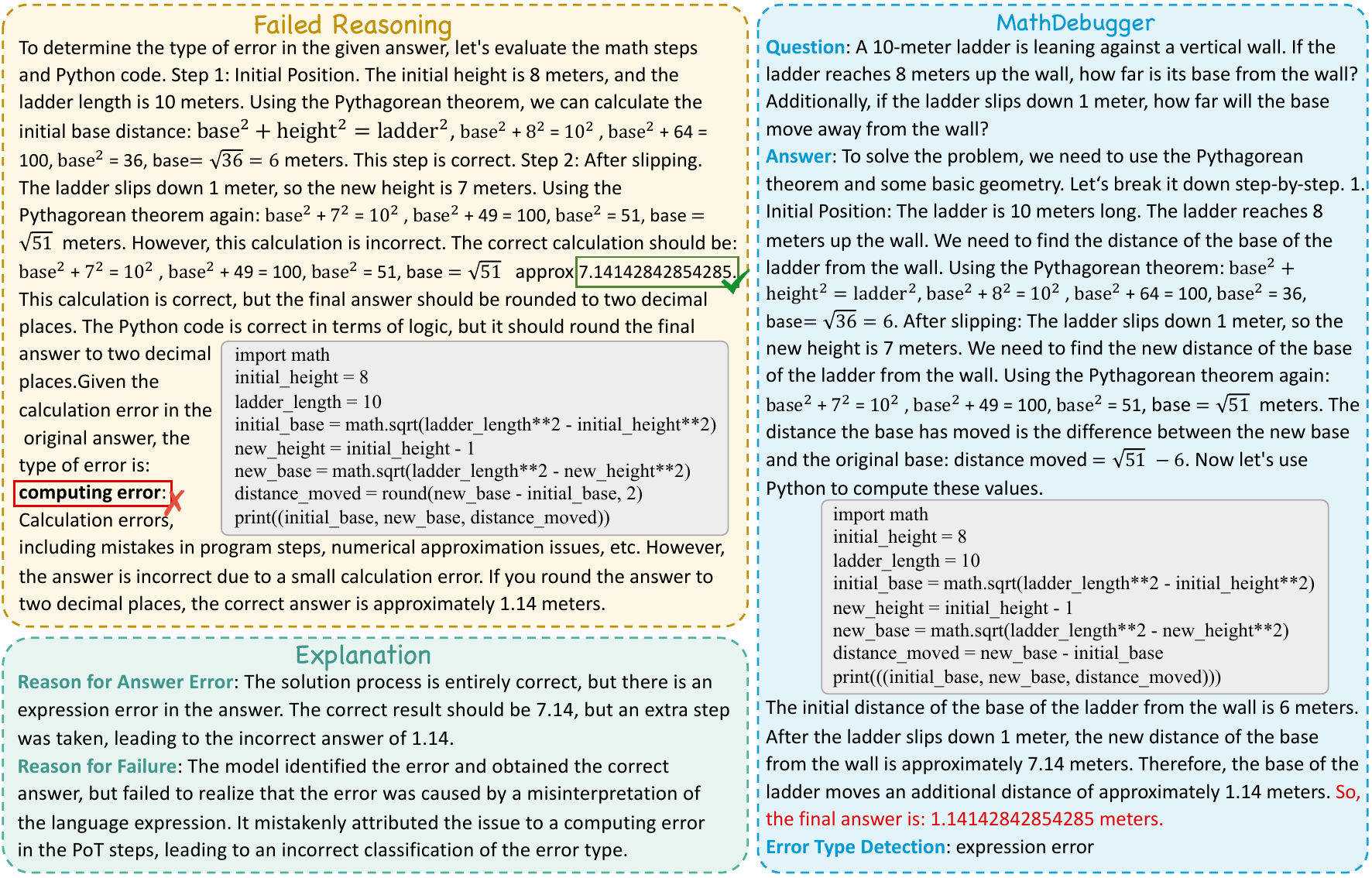}
    \caption{On a ladder problem, the model confuses a numerical error with an \textit{expression} error.}
    \label{Fig.appendix_example3}
\end{subfigure}

\vspace{6pt}

\begin{subfigure}[t]{0.82\textwidth}
    \centering
    \includegraphics[width=\linewidth]{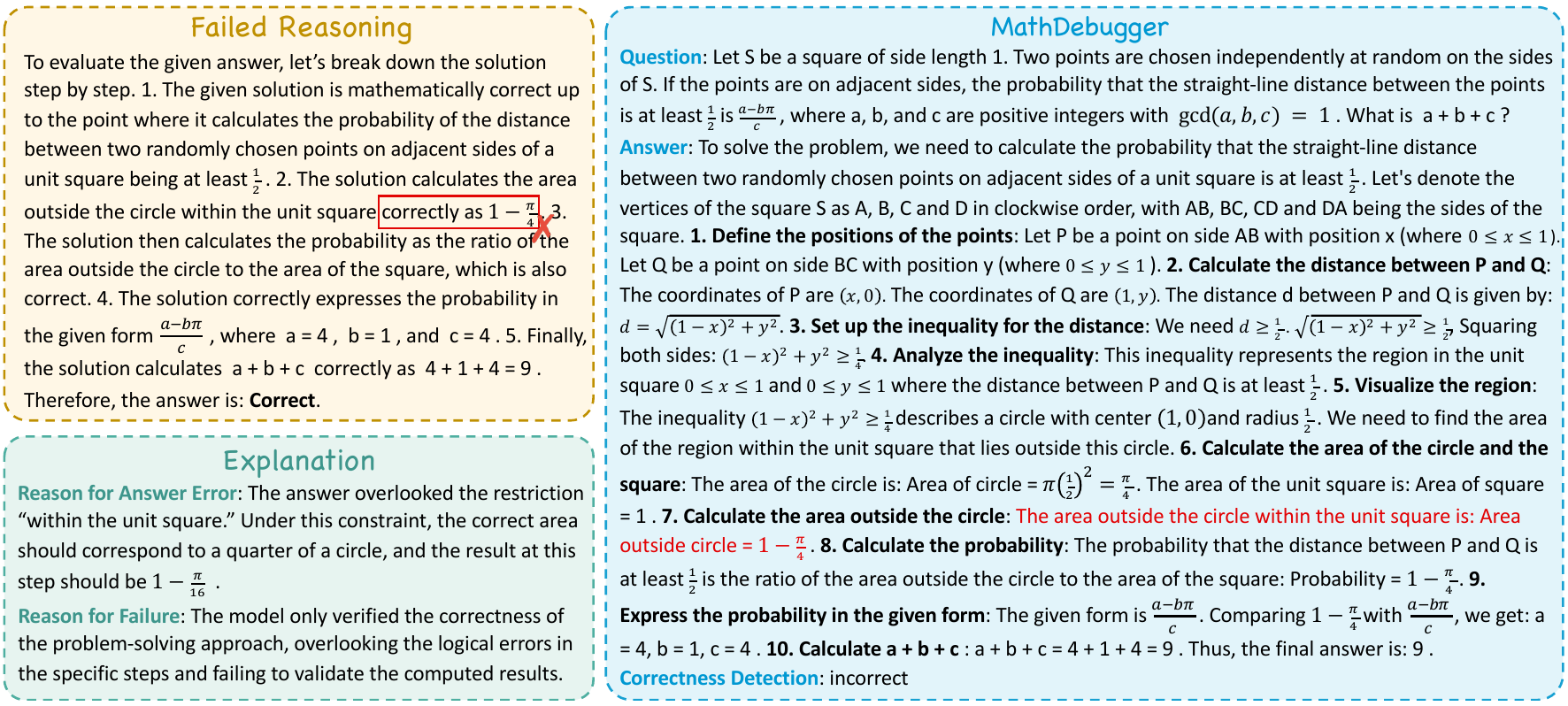}
    \caption{The model overlooks the unit-square restriction when computing the relevant circular area.}
    \label{Fig.appendix_example4}
\end{subfigure}
\caption{Additional failures involving numerical classification and geometric constraints.}
\label{fig:additional_failures_pair2}
\end{figure*}

Figure~\ref{Fig.example} in the main body shows two representative erroneous instances; \S\ref{sec:case_study} and Figure~\ref{Fig.case_study} provide a walk-through of a GPT-o1 question-level failure. Figures~\ref{Fig.appendix_example1}--\ref{Fig.appendix_example4} supply additional cases spanning both correctness and error-type detection, across question-side and answer-side instances.

\FloatBarrier

\section{Prompt Appendix}
\label{app:prompts}

\begin{figure*}[!htbp]
    \centering
    \begin{subfigure}[t]{0.90\textwidth}
        \centering
        \includegraphics[width=\linewidth]{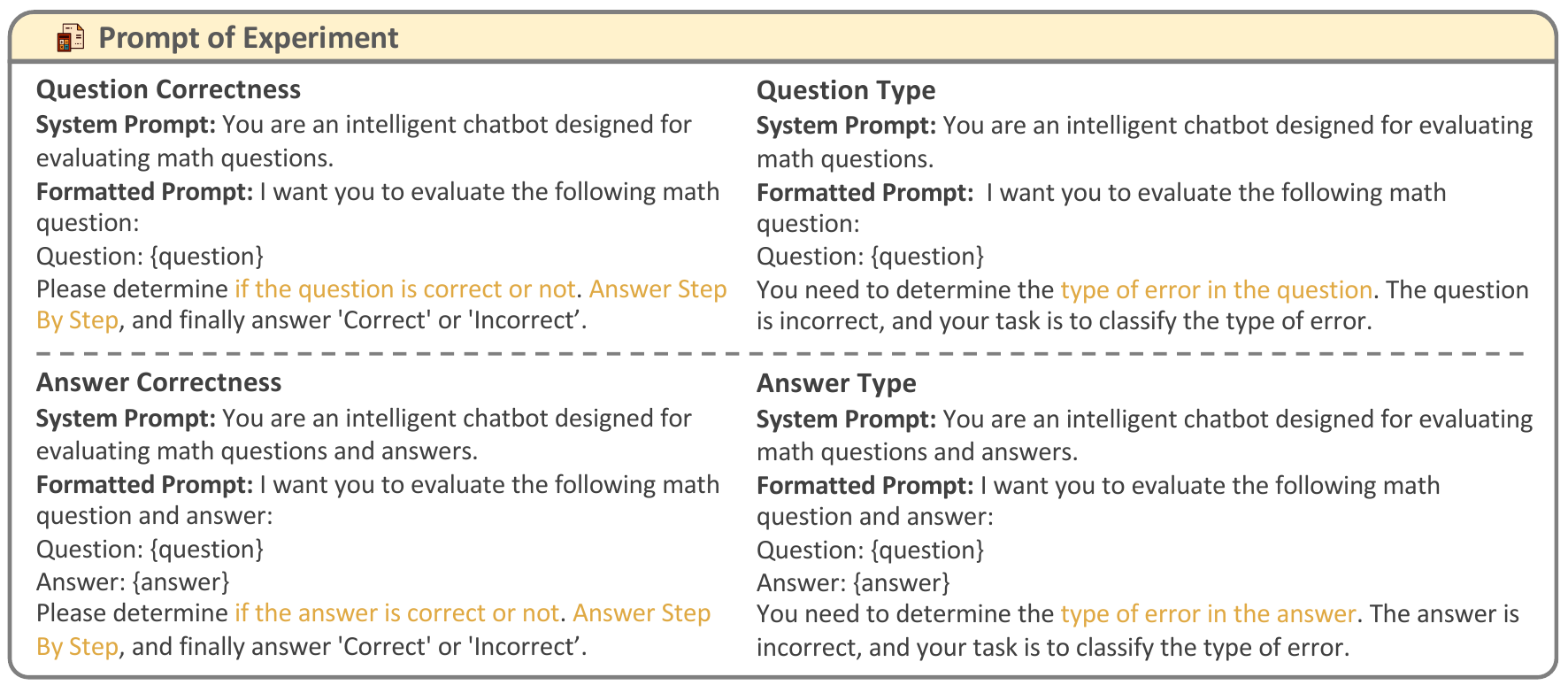}
        \caption{Experimental evaluation of correctness and error-type detection.}
        \label{Fig.experiment_prompt}
    \end{subfigure}

    \vspace{8pt}

    \begin{subfigure}[t]{0.90\textwidth}
        \centering
        \includegraphics[width=\linewidth]{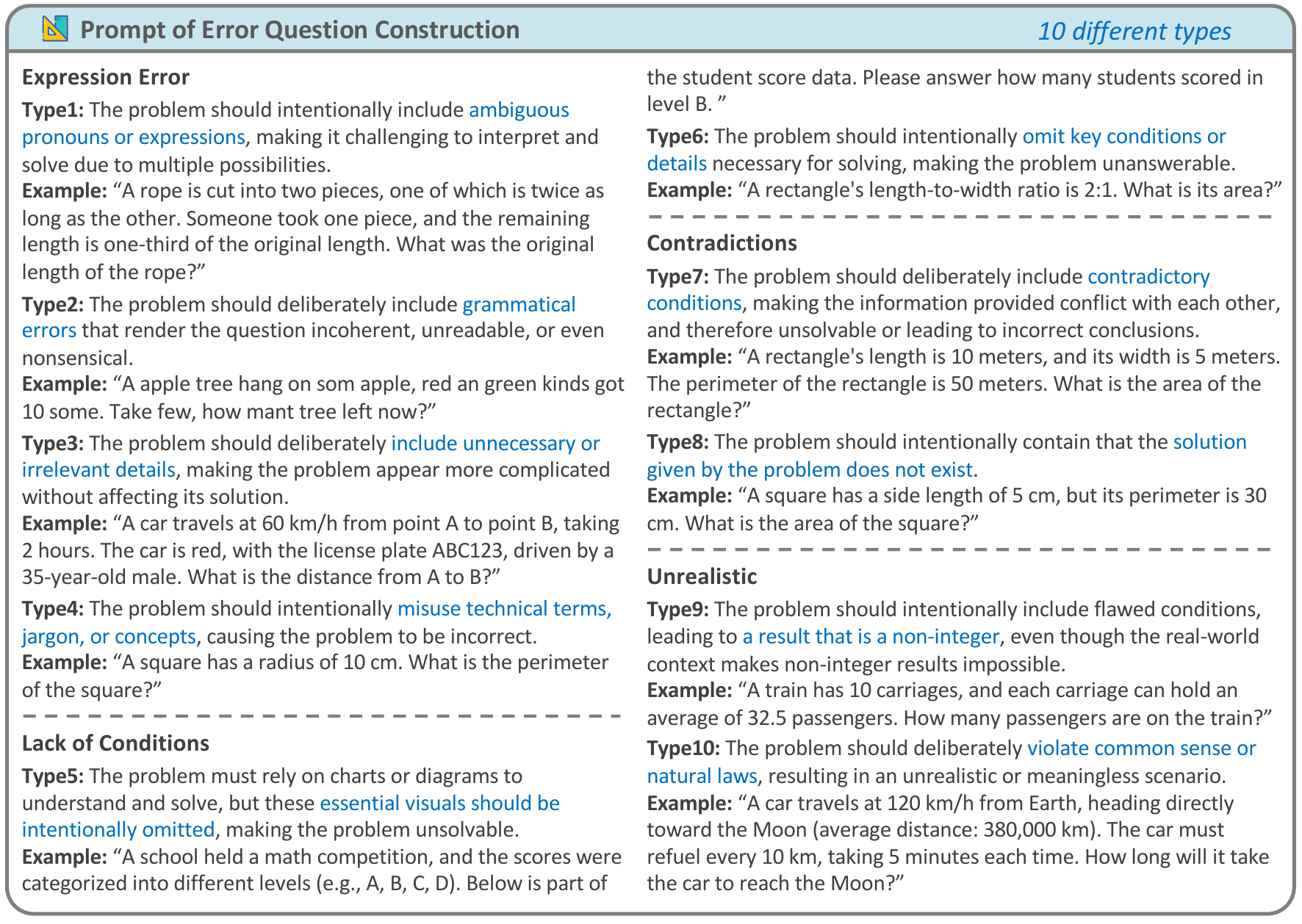}
        \caption{Error-type expansion used to construct the question dataset.}
        \label{Fig.error_question_prompt}
    \end{subfigure}
    \caption{Prompt templates for evaluation and error generation.}
    \label{fig:evaluation_and_error_prompts}
\end{figure*}

\begin{figure}[H]
    \centering
    \includegraphics[width=1.00\textwidth]{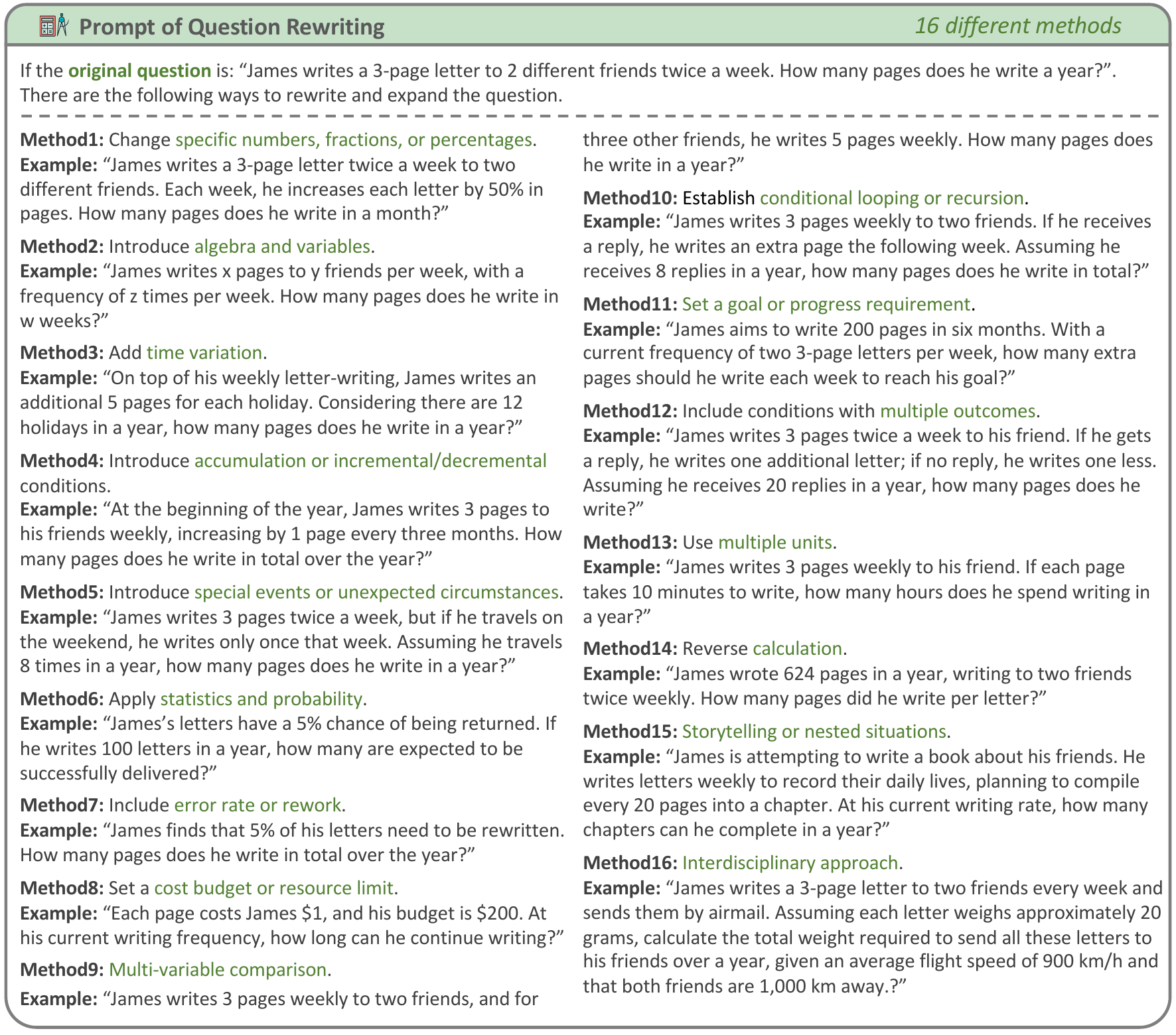}
    \caption{Question-diversification methods used to construct the answer dataset.}
    \label{Fig.question_extension_prompt}
\end{figure}

This section collects all prompts used throughout the paper: data preparation (Figure~\ref{Fig.error_question_prompt}), data synthesis (Figure~\ref{Fig.question_extension_prompt}), and the evaluation protocol (Figure~\ref{Fig.experiment_prompt}). All prompts were iteratively refined and validated during pilot experiments and can serve as references for future work on synthetic mathematical data quality.

\FloatBarrier

\end{document}